\documentclass[lettersize,journal]{IEEEtran}
\usepackage{amsmath,amsfonts}
\usepackage{algorithmic}
\usepackage{algorithm}
\usepackage{array}
\usepackage[caption=false,font=normalsize,labelfont=sf,textfont=sf]{subfig}
\usepackage{textcomp}
\usepackage{stfloats}
\usepackage{url}
\usepackage{verbatim}
\usepackage{graphicx}
\usepackage{cite}
\usepackage{xcolor}
\usepackage{multirow}
\usepackage{makecell}
\usepackage{booktabs}
\begin{document}

\title{IGEV++: Iterative Multi-range Geometry Encoding Volumes for Stereo Matching}

\author{\textbf{Gangwei Xu}, \textbf{Xianqi Wang}, \textbf{Zhaoxing Zhang}, \textbf{Junda Cheng}, \textbf{Chunyuan Liao}, \textbf{Xin Yang},~\IEEEmembership{Member,~IEEE}
\thanks{Gangwei Xu, Xianqi Wang, Zhaoxing Zhang, Junda Cheng and Xin Yang are with the School
of Electronic Information and Communications,
Huazhong University of Science and Technology, Wuhan 430074, China (E-mail:\{gwxu, xianqiw, zzx, jundacheng, xinyang2014\}@hust.edu.cn).}

\thanks{Chunyuan Liao is with HiScene Information Technology Co., Ltd., China. (E-mail: liaocy@hiscene.com).}

\thanks{Corresponding author: Xin Yang.}}

\maketitle

\begin{abstract}
Stereo matching is a core component in many computer vision and robotics systems. Despite significant advances over the last decade, handling matching ambiguities in ill-posed regions and large disparities remains an open challenge. In this paper, we propose a new deep network architecture, called IGEV++, for stereo matching. The proposed IGEV++ constructs Multi-range Geometry Encoding Volumes (MGEV), which encode coarse-grained geometry information for ill-posed regions and large disparities, while preserving fine-grained geometry information for details and small disparities. To construct MGEV, we introduce an adaptive patch matching module that efficiently and effectively computes matching costs for large disparity ranges and/or ill-posed regions. We further propose a selective geometry feature fusion module to adaptively fuse multi-range and multi-granularity geometry features in MGEV.
Then, we input the fused geometry features into ConvGRUs to iteratively update the disparity map. MGEV allows to efficiently handle large disparities and ill-posed regions, such as occlusions and textureless regions, and enjoys rapid convergence during iterations.  Our IGEV++ achieves the best performance on the Scene Flow test set across all disparity ranges, up to 768px. Our IGEV++ also achieves state-of-the-art accuracy on the Middlebury, ETH3D, KITTI 2012, and 2015 benchmarks. Specifically, IGEV++ achieves a 3.23\% 2-pixel outlier rate (Bad 2.0) on the large disparity benchmark, Middlebury, representing error reductions of 31.9\% and 54.8\% compared to RAFT-Stereo and GMStereo, respectively. We also present a real-time version of IGEV++ that achieves the best performance among all published real-time methods on the KITTI benchmarks. The code is publicly available at \textcolor{magenta}{https://github.com/gangweix/IGEV} and \textcolor{magenta}{https://github.com/gangweix/IGEV-plusplus}.
\end{abstract}


\section{Introduction}
\IEEEPARstart{S}{tereo} matching is of substantial interest because it enables the inference of 3D scene geometry from captured images, with applications spanning from 3D reconstruction to robotics and autonomous driving. The key to stereo matching is to find corresponding pixels in the left and right images. The difference between these corresponding pixel locations, known as disparity, can then be used to infer depth and reconstruct the 3D scene. 
Despite substantial efforts in the field of stereo matching~\cite{psmnet,gwcnet,raft-stereo,sttr,unistereo,croco-stereo,tosi2023nerf,tosi2024neural}, challenges persist in handling occlusions, repetitive structures, textureless or transparent objects. Additionally, effectively managing large disparities in stereo matching remains an open problem.

Deep stereo networks have become the mainstream methods with the rapid advancement of deep learning techniques and large-scale datasets.
The popular representative is PSMNet~\cite{psmnet}, which applies a 3D convolutional encoder-decoder to aggregate and regularize a 4D cost volume and uses $soft \; argmin$ to regress the disparity map from the regularized cost volume. Such cost volume filtering-based methods~\cite{gwcnet,acvnet,cfnet,pcwnet} can effectively explore stereo geometry information and achieve impressive performance on several benchmarks~\cite{kitti2012,kitti2015}. However, these methods usually construct a cost volume within a pre-defined disparity range (typically a maximum of 192px), and the final disparity prediction is derived by computing the weighted sum of those pre-defined disparity candidates. Such design greatly limits their ability in handling large disparities (up to 768px) which are widely present in high-resolution images, close-range objects, and/or wide-baseline cameras.

Constructing a full-range (i.e., image width) cost volume potentially allows for handling large disparities, but it incurs substantial computational and memory costs, limiting its application to time-constrained/hardware-constrained applications.

\begin{figure}[t]
\centering
{\includegraphics[width=1.0\linewidth]{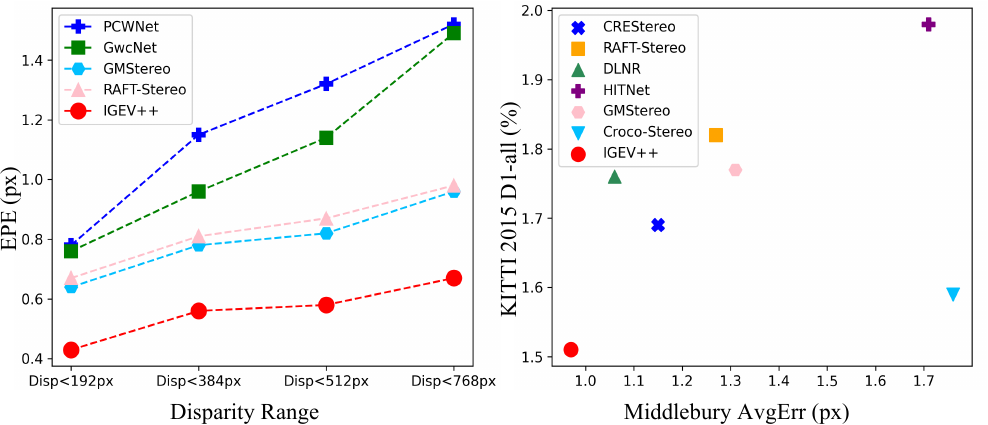}}
\vspace{-15pt}
\caption{\textbf{Left:} Comparisons with state-of-the-art stereo methods~\cite{pcwnet,gwcnet,unistereo,raft-stereo} across different disparity ranges on the Scene Flow test set~\cite{dispNetC} Our IGEV++ outperforms previously published methods by a large margin across all disparity ranges. \textbf{Right:} Comparisons with state-of-the-art stereo methods~\cite{crestereo,raft-stereo,dlnr,hitnet,unistereo,croco-stereo} on Middlebury~\cite{middlebury} and KITTI~\cite{kitti2015} leaderboards. Our IGEV++ achieves the best performance.}\label{fig:disp_range}
\vspace{-5pt}
\end{figure}

\begin{figure*}
\centering
\includegraphics[width=1.0\textwidth]{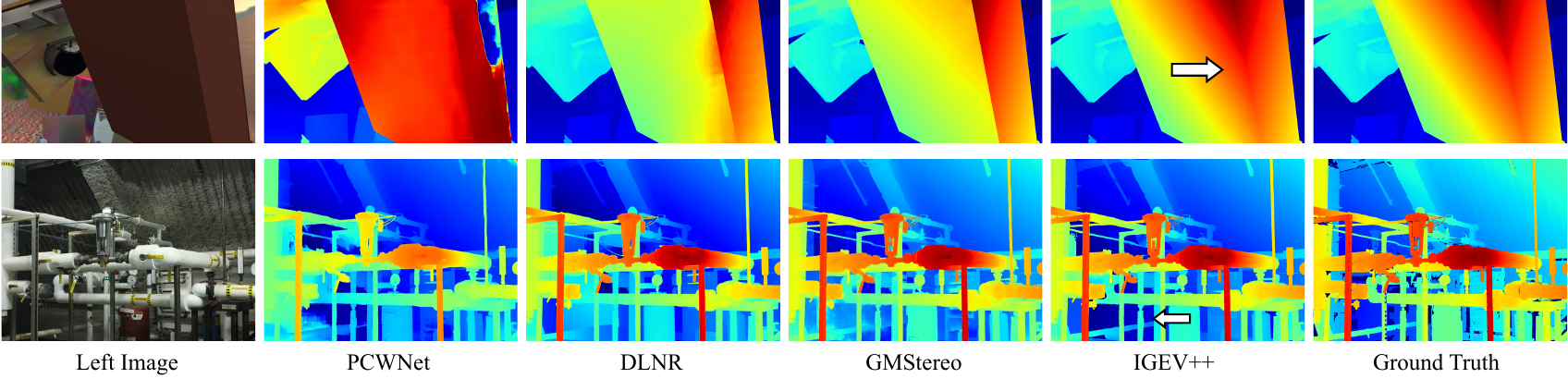}
\vspace{-15pt}
\caption{\textbf{Row 1:} Visual comparisons with state-of-the-art methods~\cite{pcwnet,dlnr,unistereo} in large disparity regions on the Scene Flow test set~\cite{dispNetC}. PCWNet~\cite{pcwnet} is a volume filtering-based method, DLNR~\cite{dlnr} is an iterative optimization-based method, and GMStereo~\cite{unistereo} is a transformer-based method. They all struggle to handle large disparities in large textureless objects at a close range. \textbf{Row 2:} Zero-shot generalization results on Middlebury~\cite{middlebury}. Our IGEV++ effectively handles large disparities in textureless regions and also distinguishes subtle details in complex backgrounds.}
\label{fig:teaser}
\vspace{-5pt}
\end{figure*}

\begin{figure}
\centering
{\includegraphics[width=1.0\linewidth]{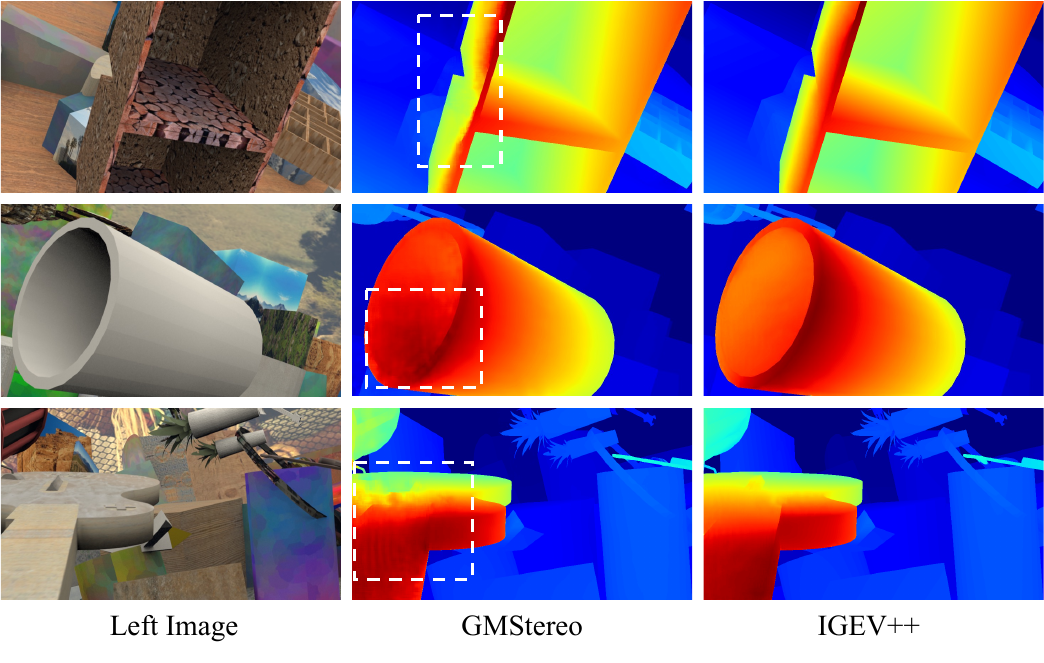}}
\caption{Comparison with the state-of-the-art transformer-based method GMStereo~\cite{unistereo} in ill-posed and large disparity regions on the Scene Flow test set.}\label{fig:sceneflow}
\vspace{-5pt}
\end{figure}

To overcome the limitations of a fixed disparity range cost volume, STTR~\cite{sttr} adopts a transformer~\cite{transformer} to compute pixel-wise correlation explicitly and densely along epipolar lines. Although STTR~\cite{sttr} has achieved promising results, it is unable to produce predictions for occluded pixels. CSTR~\cite{cstr} attempts to improve STTR’s performance with a context-enhanced transformer architecture, but it still suffers from the same limitations as STTR. GMStereo~\cite{unistereo} uses a self-attention layer to propagate predictions from non-occluded regions to occluded regions. Although the transformer-based methods can overcome the limitations of a pre-defined disparity range, they lack an explicit cost volume. As a result, they are unable to integrate multi-scale cost volumes via 3D convolutions and thus have difficulties in resolving ambiguities in ill-posed regions (Figures~\ref{fig:sceneflow},~\ref{fig:eth3d},~\ref{fig:scared}).

Recently, iterative optimization-based methods~\cite{raft-stereo,crestereo,crestereo++,dlnr, pcvnet} have exhibited attractive performance on high-resolution datasets with large disparities. Different from the filtering-based methods, iterative methods avoid the computationally expensive cost aggregation operations and progressively update the disparity map by repeatedly fetching cost information from the all-pairs 4D correlation volume. For instance, RAFT-Stereo~\cite{raft-stereo} calculates all-pairs correlations (APC) between all pixels of the left and right images on the same epipolar lines, and then exploits multi-level Convolutional Gated Recurrent Units (ConvGRUs)~\cite{gru} to recurrently update the disparity map using local costs retrieved from the APC. Due to the all-pairs 4D correlation volume, RAFT-Stereo can predict large disparities. However, without cost aggregation the original cost volume lacks non-local geometry and context information. As a result, existing iterative methods have difficulties tackling ambiguities in ill-posed regions, such as occlusions, textureless regions, and repetitive structures. Although ConvGRUs can improve the predicted disparity map by incorporating context and geometry information from context features and hidden layers, the limitation in the original cost volume greatly limits the effectiveness of each iteration, resulting in the need for a large amount of ConvGRUs iterations to achieve satisfactory performance (Figure~\ref{fig:iterations}).

In this paper, we argue that filtering-based methods and iterative optimization-based methods offer complementary advantages and limitations. The former can fully take advantage of 3D convolutions to regularize cost volumes, thereby encoding sufficient non-local geometric and contextual information into the final cost volume which is essential for disparity prediction, in particular in ill-posed regions. The latter can avoid high computational and memory costs associated with cost aggregation operations, yet are less capable in ill-posed regions based only on all-pairs correlations. 

\begin{figure*}
    \centering
    \includegraphics[width=1.0\linewidth]{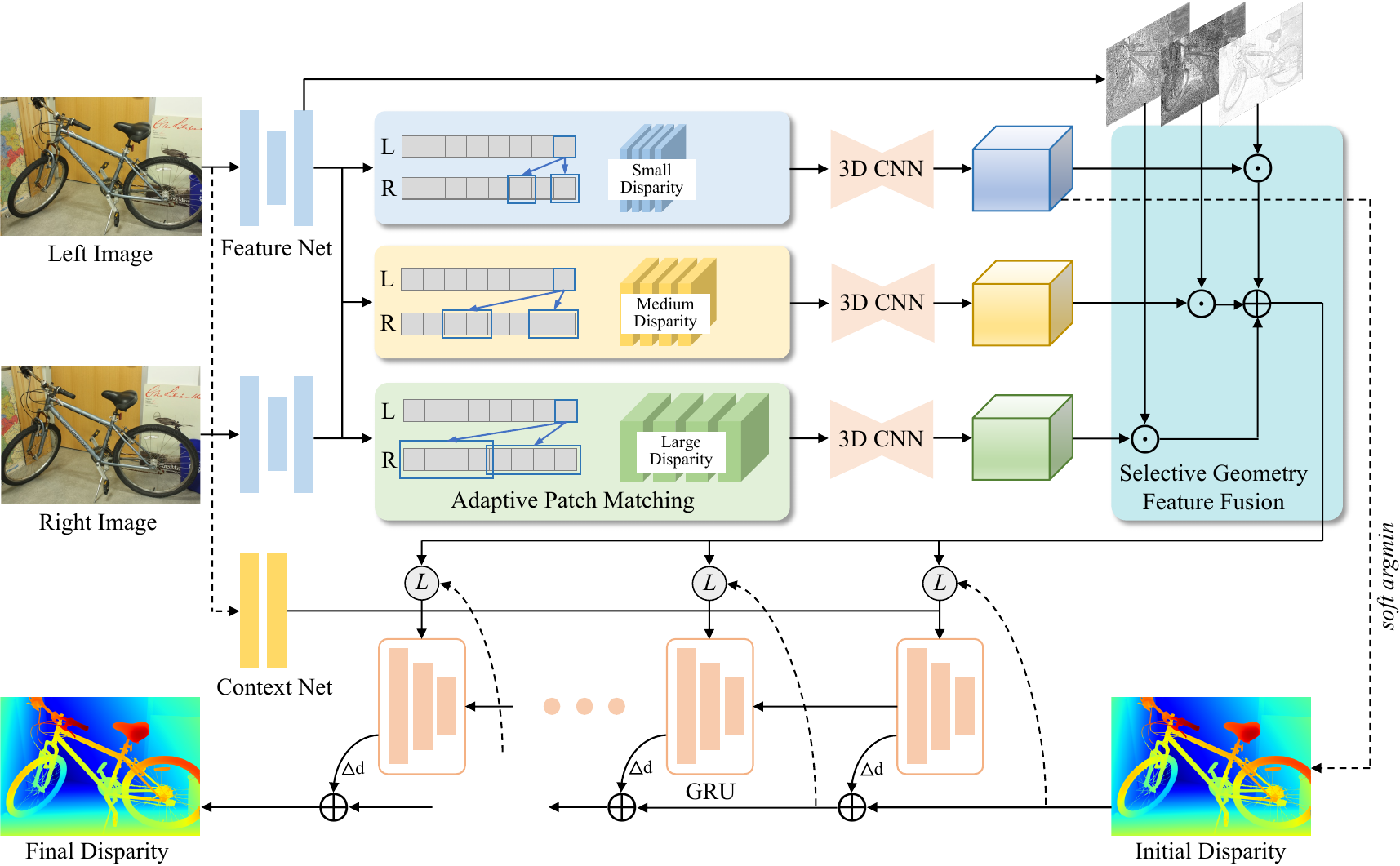}
    \caption{Network architecture of the proposed IGEV++. The IGEV++ first builds Multi-range Geometry Encoding Volumes (MGEV) via Adaptive Patch Matching (APM). MEGV encodes coarse-grained geometry information of the scene for textureless regions and large disparities and fine-grained geometry information for details and small disparities after 3D aggregation or regularization. Then we regress an initial disparity map from MGEV through $soft \; argmin$, which serves as the starting point for ConvGRUs. In each iteration, we index multi-range and multi-granularity geometry features from MGEV, selectively fuse them, and then input them into ConvGRUs to update the disparity field.}
    \label{fig:network}
\end{figure*}

To combine the complementary advantages of filtering-based and iterative optimization-based methods, we propose Iterative Geometry Encoding Volume (IGEV) and Iterative Multi-range Geometry Encoding Volumes (IGEV++), which address ambiguities in ill-posed regions by aggregating the cost volume using an extremely lightweight 3D regularization network before iterative ConvGRUs optimization. To further address the limitation of the filtering-based method and efficiently handle large disparities, our IGEV++ adopts the novel Multi-range Geometry Encoding Volumes (MGEV), which is computed effectively and efficiently via the multi-granularity matching cost calculation approach. Specifically, we compute coarse-grained matching costs (point-to-patch) for the GEV with a large disparity range, and fine-grained matching costs (point-to-point) for the GEV with a small disparity range, as illustrated in Figure~\ref{fig:network}. MGEV encodes coarse-grained geometry information for ill-posed regions and large disparities, while preserving fine-grained geometry information for detailed structures and small disparities. Our MGEV is inspired by a key observation that objects with small disparities are distant and occupy fewer pixels, while objects with large disparities are nearby and occupy more pixels. We further propose an adaptive patch matching method to enable effective and efficient MGEV construction and a selective geometry feature fusion module to efficiently integrate multi-range and multi-granularity information at each iteration.

Our IGEV++ consistently outperforms existing methods~\cite{pcwnet,raft-stereo,dlnr,unistereo} by a large margin across all disparity ranges (see Figure~\ref{fig:disp_range} and Table~\ref{tab:large_disp}). Specifically, as the disparity range increases, existing methods exhibit a significant drop in accuracy. In contrast, our method remains robust for large
disparity ranges. Our method also demonstrates outstanding ability in handling extensive ill-posed regions, achieving the best performance on reflective regions in the KITTI 2012 benchmark. 
Additionally, the proposed MGEV provides more comprehensive yet concise information for ConvGRUs to update, enabling our IGEV++ to converge faster. For example, our IGEV++ achieves a lower EPE (i.e., 0.79) with only 4 iterations (see Table~\ref{tab:iter}), compared to DLNR's 32 iterations (i.e., EPE of 0.81 for inference).

To fully exert the advantages of the proposed geometry encoding volume, we also introduce a real-time version of IGEV++, named RT-IGEV, to offer an attractive solution for time-constrained applications. Our RT-IGEV achieves real-time speed and the best accuracy among all published real-time methods~\cite{hitnet,deeppruner,coex,bgnet,fast-acv} on the KITTI benchmarks~\cite{kitti2012,kitti2015}. 

In summary, our main contributions are as follows:
\begin{itemize}
    \item We propose IGEV++, a novel deep network architecture for stereo matching, which combines the complementary advantages of filtering-based and optimization-based methods.
    \item We propose the novel Multi-range Geometry Encoding Volumes (MGEV) which encode comprehensive yet concise geometry information to enable effective disparity optimization in each iteration. Our MGEV can well resolve matching ambiguities in ill-posed regions and efficiently handle large disparities and meanwhile facilitate accurate prediction in regions with details and small disparities.
    \item We introduce an adaptive patch matching module for effective and efficient MGEV construction, and a selective geometry feature fusion module to adaptively fuse geometry features across multiple ranges and granularities.
    \item Our IGEV++ achieves state-of-the-art accuracy on four popular benchmarks: Middlebury, ETH3D, KITTI 2012, and KITTI 2015. Specifically, IGEV++ outperforms RAFT-Stereo and GMStereo by 31.9\% and 54.8\%, respectively, on the Bad 2.0 metric on the large disparity benchmark, Middlebury. Our IGEV++ also achieves the highest accuracy on the Scene Flow test set within a large disparity range of 768px (see Figure~\ref{fig:disp_range}).
    \item We propose RT-IGEV, a real-time version of IGEV++, which achieves real-time inference while delivering the best performance among all published real-time methods.
\end{itemize}

\section{Related Work}
\subsection{Cost Volume Filtering-based Methods} 
Cost Volume Filtering-based Methods typically consist of four steps, i.e. feature extraction, cost volume construction, cost aggregation and disparity regression. To improve the representative ability of a cost volume, most existing learning-based stereo methods~\cite{psmnet,nie2019multi, multilevel, segstereo, sspcv,  decomposition, fast-acv ,coatrsnet,tosi2021smd, cgi, lsp, acfnet,shi2023bidirectional,semistereo} construct a cost volume using powerful CNN features. However, the cost volume could still suffer from the ambiguity problem in occluded regions, large textureless/reflective regions and repetitive structures. The 3D convolutional networks have exhibited great potential in regularizing or filtering the cost volume, which can propagate reliable sparse matches to ambiguous and noisy regions. GCNet~\cite{gcnet} firstly uses 3D encoder-decoder architecture to regularize a 4D concatenation volume. PSMNet~\cite{psmnet} proposes a stacked hourglass 3D CNN in conjunction with intermediate supervision to regularize the concatenation volume. GwcNet~\cite{gwcnet} and ACVNet~\cite{acvnet,fast-acv} propose the group-wise correlation volume and the attention concatenation volume, respectively, to improve the expressiveness of the cost volume and in turn improve the performance in ambiguous regions. PCWNet~\cite{pcwnet} propose pyramid combination and warping cost volume to improve cross-domain generalization~\cite{dsmnet, graftnet,chang2023domain,chuah2022itsa,rao2023masked,fc}. GANet~\cite{ganet} designs a semi-global aggregation layer and a locally guided aggregation layer to further improve the accuracy. However, the high computational and memory costs of 3D CNNs often prevent these models from being applied to high-resolution cost volumes. To improve memory efficiency, several cascade methods~\cite{cfnet, cascade, fast-acv} have been proposed. CFNet~\cite{cfnet} and CasStereo~\cite{cascade} build a cost volume pyramid in a coarse-to-fine manner to progressively narrow down the predicted disparity range. Despite their impressive performance, the coarse-to-fine methods inevitably involve accumulated errors at coarse resolutions. In contrast, our IGEV++ can iteratively update disparity map at a high resolution through ConvGRUs.

\subsection{Iterative Optimization-based Methods} 
Recently, several iterative optimization-based methods~\cite{raft,raft-stereo,itermvs,crestereo,feng2024mc,dlnr,li2024local,patchmatch++} have been proposed and have achieved impressive performance in dense correspondence tasks. RAFT~\cite{raft} pioneeringly introduces a deep iterative architecture for optical flow estimation that uses ConvGRU to iteratively update the flow field with high accuracy and robustness. Subsequently, RAFT-Stereo~\cite{raft-stereo} introduces multi-level ConvGRUs to iteratively update the disparity field using local cost values retrieved from all-pairs correlations. CREStereo~\cite{crestereo} designs a hierarchical network with recurrent refinement to update the disparity field in a coarse-to-fine manner. To mitigate the loss of detail information during the iterations, DLNR~\cite{dlnr} designs the Decouple LSTM module to decouple the hidden state from the update operator. However, these iterative methods mainly rely on the all-pairs correlations (APC) which lack non-local information and thus face challenges in resolving local ambiguities in ill-posed regions. Our IGEV++ also adopts ConvGRUs as RAFT-Stereo~\cite{raft-stereo} to iteratively update the disparity map. But different from RAFT-Stereo~\cite{raft-stereo}, we construct MGEV to encode non-local geometry and context information which can significantly improve the effectiveness of each iteration. In addition, we provide an accurate initial disparity map for the ConvGRUs updater to start at a negligible cost, yielding a much faster convergence than RAFT-Stereo~\cite{raft-stereo}.

\subsection{Transformer-based Stereo Methods}
With the rise of attention mechanisms, transformer-based deep stereo methods~\cite{sttr,cstr,elfnet,unistereo} have become another line of stereo matching research. A representative method is STTR~\cite{sttr}, which revisits the stereo matching problem from a sequence-to-sequence correspondence perspective with transformer to replace the fixed-range cost volume. Following this, CSTR~\cite{cstr} introduces a context-enhanced transformer architecture to improve STTR's performance. GMStereo~\cite{unistereo} leverages the transformer to integrate knowledge between the left and right images via cross-view interactions. Recently, Croco-Stereo~\cite{croco-stereo} has achieved impressive performance by scaling up vision transformer-based cross-completion architectures with large amounts of data. While transformer-based stereo methods excel in modeling long-range dependencies and can handle large disparities to some extent, they encounter difficulties in addressing local ambiguities in ill-posed regions (such as occlusions and reflective regions) due to the absence of an explicit cost volume.

\section{Method}
\label{sec:method}
In this section, we detail the network architecture of IGEV++, which consists of multi-scale feature extraction, multi-range geometry encoding volumes construction, and ConvGRU-based update operator.

\subsection{Multi-scale Feature Extraction}

Feature extraction contains two parts: 1) a feature network that extracts multi-scale features for cost volume construction and cost aggregation guidance, and 2) a context network that extracts multi-scale context features for ConvGRUs hidden state initialization and update.

\textbf{Feature Network.} Given the left and the right images $\mathbf{I}_{l(r)}\in\mathbb{R}^{{3}\times{H}\times{W}}$, we first apply the MobileNetV2 pretrained on ImageNet~\cite{imagenet} to scale $\mathbf{I}_{l(r)}$ down to 1/32 of the original size, and then use upsampling blocks with skip-connections to recover them up to 1/4 scale, resulting in multi-scale features \{$\mathbf{f}_{l,i}(\mathbf{f}_{r,i})\in\mathbb{R}^{{C_i}\times{\frac{H}{i}}\times{\frac{W}{i}}}$\} ($i$=4, 8, 16, 32 and $C_i$ for feature channels). The $\mathbf{f}_{l,4}$ and $\mathbf{f}_{r,4}$ are used to construct the multi-range cost volumes. And the $\mathbf{f}_{l,i}$ ($i$=4, 8, 16, 32) are also used as context guidance for cost volume filtering.

\textbf{Context Network.} Following RAFT-Stereo~\cite{raft-stereo}, the context network consists of a series of residual blocks~\cite{resnet} and downsampling layers, producing multi-scale context features at 1/4, 1/8 and 1/16 of the input image resolution with 128 channels. The multi-scale context features are used to initialize the hidden state of the ConvGRUs and are also inserted into the ConvGRUs to update the disparity map at each iteration.

\subsection{Multi-range Geometry Encoding Volumes Construction}
Given the left feature map $\mathbf{f}_{l,4}$ and right feature map $\mathbf{f}_{r,4}$ extracted from $\mathbf{I}_{l}$ and $\mathbf{I}_{r}$, we construct group-wise correlation volumes respectively for small disparity range $D^s$ (\textless 192px), medium disparity range $D^m$ (\textless 384px), and large disparity range $D^l$ (\textless 768px). The group-wise correlation volume~\cite{gwcnet} refers to splitting features $\mathbf{f}_{l,4}$ ($\mathbf{f}_{r,4}$) into $N_g$ ($N_g$=8) groups along the channel dimension and computes correlation maps group by group. For small disparity ranges, the correlation volume $\mathbf{C}^{s}$ is constructed by,
\begin{equation}
\mathbf{C}^{s}(g,d^s,x,y)=\frac{1}{N_c/N_g}\langle\mathbf{f}_{l,4}^g(x,y), \mathbf{f}_{r,4}^g(x-d^s,y)\rangle, 
\end{equation}

where $\langle\cdot,\cdot\rangle$ is the inner product, $d^s$ is the disparity index, $d^s\in\mathcal{D}^s$, $\mathcal{D}^s=\{0,1,2,\dots,D^s/4-1\}$, $N_c$ denotes the number of feature channels. 

However, $\mathbf{C}^{s}$ measures only a limited disparity range. Simply increasing $D^s$ would significantly increase the computation and memory overhead of cost aggregation or regularization, limiting its application to high-resolution images. To efficiently handle large disparities, we construct medium-range correlation volume $\mathbf{C}^{m}$ and large-range correlation volume $\mathbf{C}^{l}$ that encode coarse-grained matching information using \textbf{Adaptive Patch Matching}. The construction process of $\mathbf{C}^{m}$ and $\mathbf{C}^{l}$ is similar. For brevity, we only detail the construction process of $\mathbf{C}^{l}$ here. Specifically, the correlation volume $\mathbf{C}^{l}$ is constructed by,

\begin{equation}
\begin{aligned}
\mathbf{C}^{l}(g,d^l,x,y)\!= & \; \!\frac{1}{N_c/N_g}\langle\mathbf{f}_{l,4}^g(x,y), \mathbf{p}^{g}(x-d^l,y)\rangle, \\
\mathbf{p}^g(x-d^l,y) = & \; \sum\limits_{i=0}^{3}\omega_{i}\mathbf{f}_{r,4}^g(x-(d^l+i),y)
\end{aligned}
\end{equation}

where the candidate disparity values $d^l\in\mathcal{D}^l$, $\mathcal{D}^l=\{0,4,8,\dots, D^l/4-4\}$, consisting of 48 disparity candidates (768/4/4) for cost aggregation. $\mathbf{C}^{l}$ is constructed by point-to-patch matching (coarse-grained matching) between the left and right images (Figure~\ref{fig:network}). $\omega_{i}$ represents the weight of the matching cost in a patch of the right image and is learned adaptively during the training process. To efficiently handle large disparities, we propose adaptive patch matching to represent a large disparity range with fewer disparity candidates, significantly reducing the computational cost of cost aggregation and lowering the difficulty of disparity regression.

Constructing $\mathbf{C}^{s}$ and $\mathbf{C}^{l}$ based on only feature correlations usually encounters matching noises and ambiguity, and lacks the capability to capture global geometric structures and spatial evidence. To address this problem, we further process $\mathbf{C}^{s}$, $\mathbf{C}^{m}$ and $\mathbf{C}^{l}$ using a lightweight 3D regularization network $\mathbf{R}$ to obtain the multi-range geometry encoding volumes $\mathbf{G}^{s}$,  $\mathbf{G}^{m}$, and $\mathbf{G}^{l}$. Taking $\mathbf{C}^{l}$ as an example, $\mathbf{G}^{l}$ is obtained through,
\begin{equation}
\mathbf{G}^{l}=\mathbf{R}(\mathbf{C}^{l})
\end{equation}

The 3D regularization network $\mathbf{R}$ is based on a lightweight 3D UNet that consists of three down-sampling blocks and three up-sampling blocks. Each down-sampling block consists of two $3\times3\times\!3$ 3D convolutions. The number of channels of the three down-sampling blocks is 16, 32, and 48 respectively. Each up-sampling block consists of a $4\times4\times4$ 3D transposed convolution and two $3\times3\times3$ 3D convolutions.
We follow CoEx~\cite{coex}, which excites the cost volume channels with weights computed from the left features for cost aggregation. For a $\frac{D}{i}\times\frac{H}{i}\times\frac{W}{i}$ cost volume $\mathbf{C}_i$ ($i$=4, 8, 16 and 32) in cost aggregation, the guided cost volume excitation is expressed as,
\begin{equation}
\mathbf{C}_{i}^{'}=\sigma(\mathbf{f}_{l,i})\odot\mathbf{C}_i ,
\end{equation}
where $\sigma$ is the sigmoid function, $\odot$ denotes the Hadamard Product. The 3D regularization network, which inserts guided cost volume excitation operation, can effectively infer and propagate scene geometry information at multiple scales, leading to the MGEV. The MGEV encodes multi-range and multi-granularity geometric information which provides more comprehensive yet concise information for ConvGRUs optimization and in turn helps eliminating matching noises and resolving matching ambiguities in ill-posed regions.

\begin{figure}
\centering
{\includegraphics[width=0.9\linewidth]{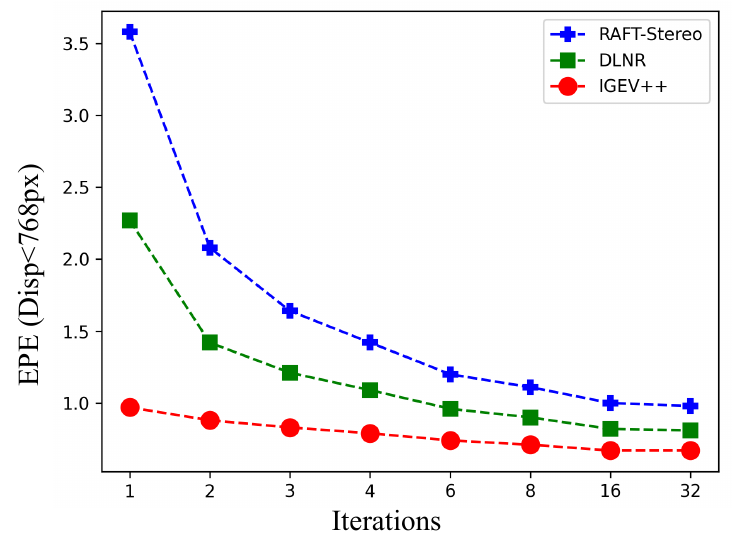}}
\caption{\textbf{EPE (Disp\textless768px) vs. number of iterations at inference.} The figure exhibits the prediction results on the Scene Flow test set at different iteration numbers during inference. Our IGEV++ converges faster and reaches a lower convergence point.}\label{fig:iterations}
\vspace{-5pt}
\end{figure}

\subsection{ConvGRU-based Update Operator}
The objective of the update operator is to iteratively refine the initial disparity map using the indexed geometry features from MGEV. Before starting the iteration, we first apply $soft \; argmin$ to $\mathbf{G}^s$, $\mathbf{G}^m$, and $\mathbf{G}^l$ to regress a set of initial disparity maps $\mathbf{d}_{0}^{s}$, $\mathbf{d}_{0}^{m}$, and $\mathbf{d}_{0}^{l}$,

\begin{equation}
\begin{aligned}
\mathbf{d}_{0}^s = & \; \sum\limits_{d^s\in\mathcal{D}^s} d^s \times Softmax(\mathbf{G}^s(d^s)), \\
\mathbf{d}_{0}^m = & \; \sum\limits_{d^m\in\mathcal{D}^m} d^m \times Softmax(\mathbf{G}^m(d^m)), \\
\mathbf{d}_{0}^l = & \; \sum\limits_{d^l\in\mathcal{D}^l} d^l \times Softmax(\mathbf{G}^l(d^l)). \\
\end{aligned}
\label{equ:soft}
\end{equation}

To speed up convergence, we take the initial disparity map $\mathbf{d}^{s}_{0}$ as the starting point for ConvGRU-based iterations. In each iteration, we utilize the previous disparity map to index a set of geometry features $\mathbf{f}_{G}^{s}$, $\mathbf{f}_{G}^{m}$, and $\mathbf{f}_{G}^{l}$ from $\mathbf{G}^s$, $\mathbf{G}^m$, and $\mathbf{G}^l$. We further propose a Selective Geometry Feature Fusion module (SGFF) to effectively integrate geometry features from different ranges and granularities of MGEV.

\textbf{Selective Geometry Feature Fusion}. 
Large disparity regions and ill-posed regions depend more on long-range and coarse-grained geometry features $\mathbf{f}_{G}^{l}$, while small disparity regions and detailed regions rely more on short-range and fine-grained geometry features $\mathbf{f}_{G}^{s}$. Therefore, we selectively fuse geometry features in different regions through SGFF. The initial disparity maps $\mathbf{d}_0^s$,  $\mathbf{d}_0^m$, and $\mathbf{d}_0^l$ indicate the range of disparity, while the left feature map $\mathbf{f}_{l,4}$ denotes the type of region (i.e., textureless or detailed regions). Therefore, SGFF takes the initial disparity maps $\mathbf{d}_0^s$,  $\mathbf{d}_0^m$, and $\mathbf{d}_0^l$, along with the left feature map $\mathbf{f}_{l,4}$, as inputs to predict a set of selective weights using two convolutional layers,
\begin{equation}
\begin{aligned}
\mathbf{f}_{d} = & \; \text{Conv}(\text{Concat}\{\mathbf{d}_0^s, \mathbf{d}_0^m, \mathbf{d}_0^l\}), \\
\mathbf{s}_{j} = & \; \sigma(\text{Conv}(\text{Concat}\{\mathbf{f}_{l,4}, \mathbf{f}_{d}\})), j = s, m, l. \\
\end{aligned}
\end{equation}

Based on these selective weights $\mathbf{s}_{s}$, $\mathbf{s}_{m}$, and $\mathbf{s}_{l}$, we adaptively fuse geometry features by,
\begin{equation}
\mathbf{f}_{G}=\mathbf{s}_{s}\odot\mathbf{f}_G^s+\mathbf{s}_{m}\odot\mathbf{f}_G^m+\mathbf{s}_{l}\odot\mathbf{f}_G^l.
\end{equation}

In addition to the fused geometry feature $\mathbf{f}_{G}$, we also index correlation values from the all-pairs correlation volume and concatenate them with $\mathbf{f}_{G}$.

In each iteration, the fused geometry feature $\mathbf{f}_{G}$ and the previous disparity map $\mathbf{d}_{k-1}$ are passed through two encoder layers and then concatenated with $\mathbf{d}_{k-1}$ to form $x_k$. Then we use ConvGRUs to update the hidden state $h_{k-1}$ as RAFT-Stereo~\cite{raft-stereo},
\begin{equation}
\begin{aligned}
x_k = & \; [\text{Encoder}_{g}(\mathbf{f}_G), \text{Encoder}_{d}(\mathbf{d}_{k-1}),\mathbf{d}_{k-1}]\\
z_k = & \;\sigma(\text{Conv}([h_{k-1}, x_k], W_z) + c_z), \\
r_k = & \;\sigma(\text{Conv}([h_{k-1}, x_k], W_r) + c_r), \\
\Tilde{h}_k = & \,\tanh(\text{Conv}([r_k \odot h_{k-1}, x_k], W_h) + c_h), \\
h_k = & \;(1-z_k) \odot h_{k-1} + z_k \odot \Tilde{h}_k,
\end{aligned}
\end{equation}
where $W_z$, $W_r$, and $W_h$ are the parameters of the network, $c_z$, $c_r$, $c_h$ are context features generated from the context network. The number of channels for both the hidden states and the context features is 128. The $\text{Encoder}_{g}$ and $\text{Encoder}_{d}$ consist of two convolutional layers respectively. Based on the hidden state $h_k$, we decode a residual disparity $\triangle \mathbf{d}_k$ through two convolutional layers, then we update the disparity map,
\begin{equation}
\begin{aligned}
\mathbf{d}_{k} = & \; \mathbf{d}_{k-1} + \triangle \mathbf{d}_k
\end{aligned}
\end{equation}

\begin{table*}
\centering
\caption{Comparisons with Existing Representative Methods on the Scene Flow test set across multiple disparity ranges. Disp\textless192px, 384px, 512px, and 768px indicate that the method is evaluated in regions where the ground-truth disparity $\mathbf{d}_{gt}$ is less than 192px, 384px, 512px, and 768px, respectively.} \label{tab:large_disp}
\begin{tabular}{l|cc|cc|cc|ccc}
\toprule
\multirow{2}{*}{Method} &\multicolumn{2}{c|}{Disp\textless192px}  &\multicolumn{2}{c|}{Disp\textless384px} &\multicolumn{2}{c|}{Disp\textless512px} &\multicolumn{3}{c}{Disp\textless768px}  \\ 
& EPE &Bad 3.0 & EPE &Bad 3.0 & EPE &Bad 3.0 & EPE &Bad 3.0 & Memory (G) \\ 
\midrule
PCWNet~\cite{pcwnet} & 0.78 & 3.29 & 1.15 & 3.77 &1.32 &3.87 & 1.52 & 4.00 & 11.62  \\
GwcNet~\cite{gwcnet} & 0.76 & 3.30 & 0.96 & 3.61 &1.14 & 3.69 & 1.49 &3.78  & 10.04 \\
RAFT-Stereo~\cite{raft-stereo} &0.67 &3.41 &0.81 &3.73 & 0.87 & 3.79 & 0.98 & 3.83 & 1.65 \\
GMStereo~\cite{unistereo} & 0.64 & 2.55 & 0.78 & 2.85 & 0.82 & 2.87 & 0.96 & 2.90 & 1.56 \\
IGEV++ (Ours) & \textbf{0.43} & \textbf{1.81} & \textbf{0.56} & \textbf{2.10} & \textbf{0.58} & \textbf{2.16} & \textbf{0.67} & \textbf{2.21} & 1.57 \\
\bottomrule
\end{tabular}
\end{table*}

\begin{figure}
\centering
{\includegraphics[width=1.0\linewidth]{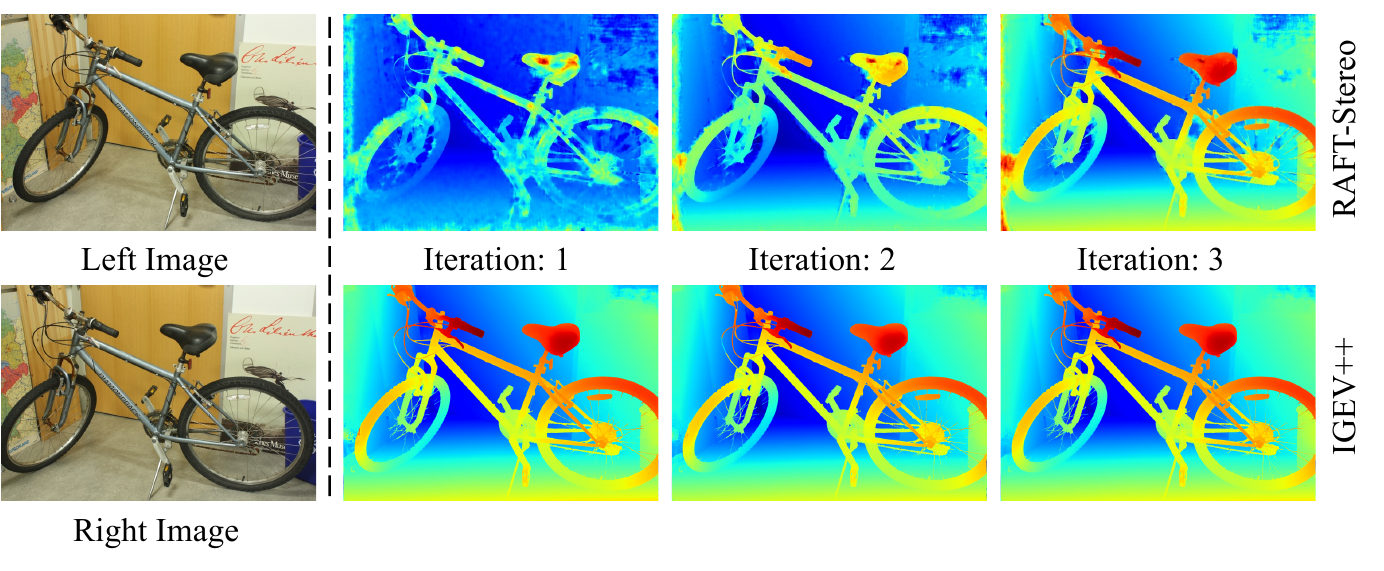}}
\vspace{-10pt}
\caption{Visual comparisons with RAFT-Stereo~\cite{raft-stereo} using few iterations for inference, with iteration numbers of 1, 2, and 3.}\label{fig:iter123}
\end{figure}

\subsection{Spatial Upsampling}
We output a full-resolution disparity map by the weighted combination of the predicted disparity $\mathbf{d}_k$ at 1/4 resolution. Different from RAFT-Stereo~\cite{raft-stereo} which predicts weights from the hidden state ${h_k}$ at 1/4 resolution, we utilize the higher resolution context features to obtain the weights. We convolve the hidden state to generate features and then upsample them to 1/2 resolution. The upsampled features are concatenated with $\mathbf{f}_{l,2}$ from left image to produce weights $\mathbf{W}\in\mathbb{R}^{H\times{W}\times9}$. We output the full-resolution disparity map by the weighted combination of the local neighboring points of the low-resolution disparity map $\mathbf{d}_{k}$.

\subsection{Loss Function}
The total loss consists of regularization loss and iteration loss. We compute the regularization loss $\mathcal{L}_{reg}$ between the regressed disparity maps $\mathbf{d}_0^{s}$, $\mathbf{d}_0^{m}$, and $\mathbf{d}_0^{l}$ from MGEV and ground-truth disparity map $\mathbf{d}_{gt}$,
\begin{equation}
\begin{aligned}
\mathcal{L}_{reg}^s = & \; Smooth_{L_1}(\mathbf{d}_0^s-\mathbf{d}_{gt}),\\
\mathcal{L}_{reg}^m = & \; Smooth_{L_1}(\mathbf{d}_0^m-\mathbf{d}_{gt}),\\
\mathcal{L}_{reg}^l = & \; Smooth_{L_1}(\mathbf{d}_0^l-\mathbf{d}_{gt}),\\
\mathcal{L}_{reg} = & \; \lambda_{s}\mathcal{L}_{reg}^s + \lambda_{m}\mathcal{L}_{reg}^m + \lambda_{l}\mathcal{L}_{reg}^l.
\end{aligned}
\end{equation}

We then compute the iteration loss between all updated disparity maps $\{\mathbf{d}_{i}\}_{i=1}^{N}$ and ground-truth disparity map $\mathbf{d}_{gt}$,
\begin{equation}
    \mathcal{L}_{iter} = \sum_{i=1}^{N} \gamma^{N-i} ||\mathbf{d}_i-\mathbf{d}_{gt}||_1.
\end{equation}

Finally, the total loss $\mathcal{L}_{total}$ is represented as:
\begin{equation}
    \mathcal{L}_{total} = \mathcal{L}_{reg} + \mathcal{L}_{iter}.
\end{equation}

\begin{figure*}
\centering
\includegraphics[width=0.9\textwidth]{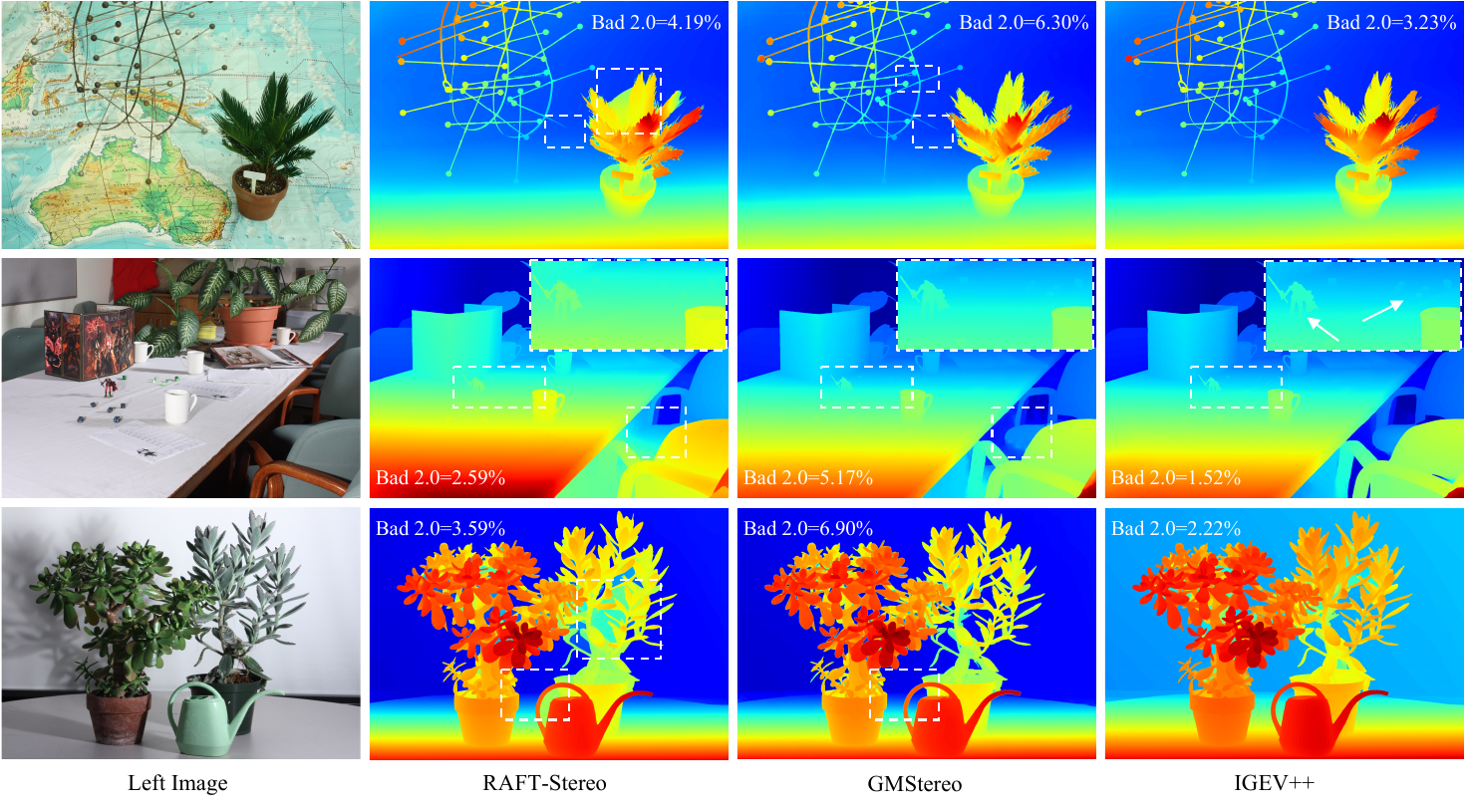} 
\vspace{-5pt}
\caption{Visual comparisons with state-of-the-art stereo methods~\cite{raft-stereo,unistereo} on the Middlebury test set. Our MGEV encodes both coarse-grained and fine-grained geometry information, thus IGEV++ can concurrently estimate large disparities in nearby textureless regions accurately, distinguish subtle details (Row 2), and predict sharp edges (Row 3, white dashed box).}
\label{fig:middlebury}
\end{figure*}

\begin{table*}
  \centering
  \caption{
  Ablation study of proposed modules on the Scene Flow test set. GEV denotes single-range geometry encoding volume, MGEV denotes multi-range geometry encoding volumes, APM denotes adaptive patch matching, and SGFF denotes selective geometry feature fusion. The baseline is RAFT-Stereo~\cite{raft-stereo} using MobileNetV2 100 as the backbone.}
  \begin{tabular}{l|cccc|cc|cc|cc|cc}
    \toprule
    \multirow{2}{*}{Model}  & \multirow{2}{*}{GEV} & \multirow{2}{*}{MGEV} & \multirow{2}{*}{APM} & \multirow{2}{*}{SGFF}  & \multicolumn{2}{c|}{Disp\textless192px}  &\multicolumn{2}{c|}{Disp\textless384px} &\multicolumn{2}{c|}{Disp\textless512px} &\multicolumn{2}{c}{Disp\textless768px}  \\
    & & & & & EPE &Bad 3.0 & EPE &Bad 3.0 & EPE &Bad 3.0 & EPE &Bad 3.0 \\
    \midrule

    Baseline & & & & & 0.54 & 2.30 & 0.68 & 2.60 & 0.74 & 2.70 & 0.87 & 2.75 \\
    IGEV &\checkmark & & & & 0.46 & 1.96 & 0.59 & 2.32 & 0.68 & 2.38 & 0.84 & 2.46 \\
    MGEV & &\checkmark & & & 0.46 & 2.00 & 0.59 & 2.29 & 0.63 & 2.31 & 0.74 & 2.35 \\
    MGEV+APM & &\checkmark &\checkmark & & 0.45 & 1.91 & 0.58 & 2.22 & 0.61 & 2.28 & 0.71 & 2.33 \\
    IGEV++ & &\checkmark &\checkmark &\checkmark & \textbf{0.43} & \textbf{1.81} & \textbf{0.56} & \textbf{2.10} & \textbf{0.58} & \textbf{2.16} & \textbf{0.67} & \textbf{2.21} \\
    \bottomrule
  \end{tabular}
  \label{tab:ablation}
\end{table*}

\begin{table*}
  \centering
  \caption{
  Quantitative comparisons between the predicted disparity maps for different ranges of geometry encoding volumes (GEV). The iterative small-range GEV performs well for small disparities but struggles with large disparities. On the other hand, the iterative large-range GEV handles large disparities effectively.}
  \begin{tabular}{l|ccc|ccc}
    \toprule
    \multirow{2}{*}{Model}  & small-range & medium-range & large-range & \multicolumn{3}{c}{Scene Flow EPE} \\
    & GEV & GEV & GEV & Disp\textless192px & Disp\textless384px & Disp\textless768px \\
    \midrule

    small-range GEV &\checkmark & & &  0.47 & 0.87 & 1.06 \\
    medium-range GEV & & \checkmark & & 0.51 & 0.67 & 0.82 \\
    large-range GEV & & & \checkmark & 0.54 & 0.69 & 0.76 \\
    multi-range GEV (final) & \checkmark &\checkmark &\checkmark & \textbf{0.43} & \textbf{0.56} & \textbf{0.67} \\
    \bottomrule
  \end{tabular}
  \label{tab:ablation_sml}
\end{table*}

\begin{figure*}
\centering
\includegraphics[width=0.9\textwidth]{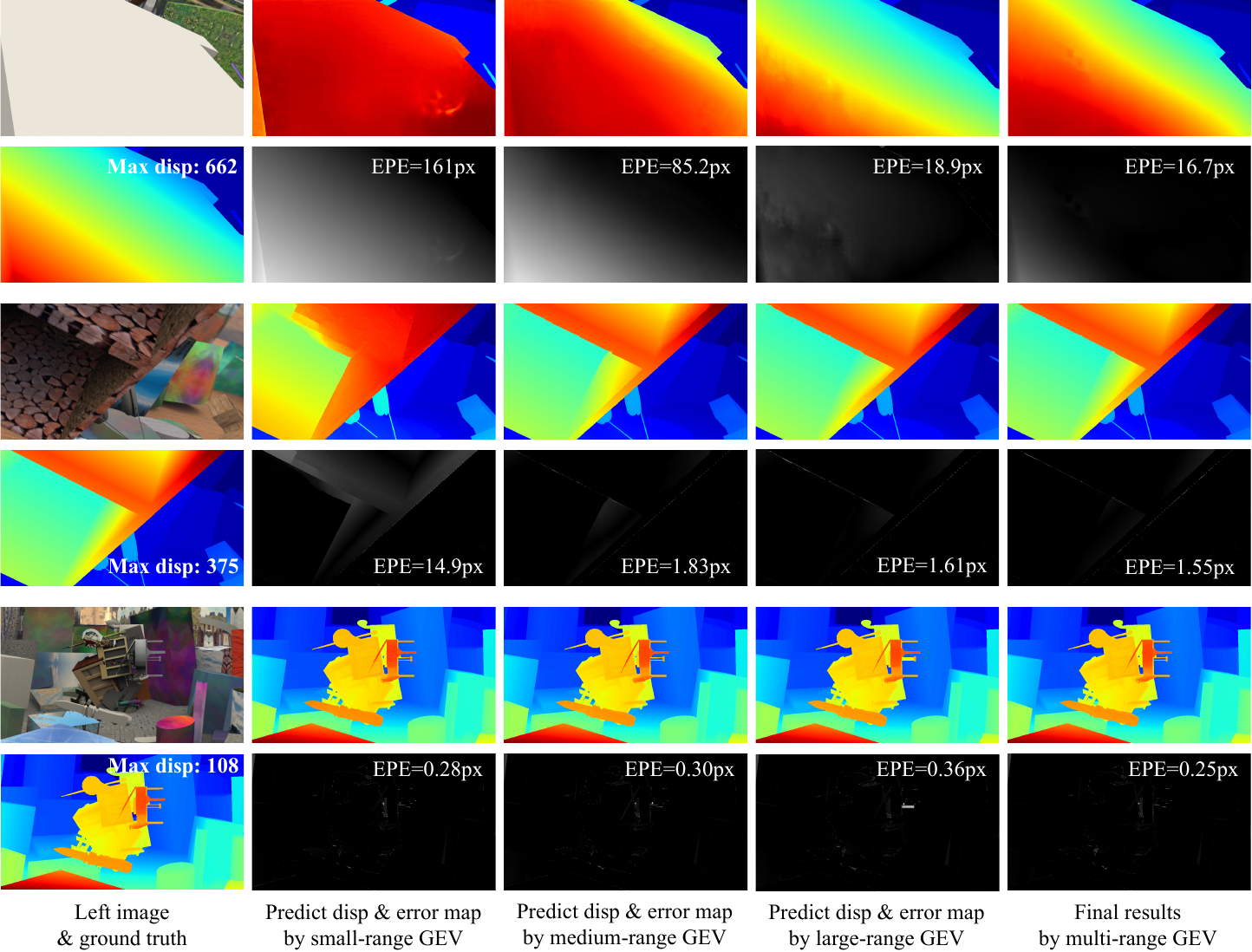} 
\caption{Qualitative comparisons between the predicted disparity maps for different ranges of geometry encoding volumes (GEV). From top to bottom, the three inputs contain different maximum disparities: 662, 375, and 108, respectively. The second, third, and fourth columns show the predicted disparity maps and error maps generated by iterating over the small-range GEV, medium-range GEV, and large-range GEV, respectively. The final column shows the results of our final MGEV.}
\label{fig:comp_sml}
\vspace{-5pt}
\end{figure*}

\section{Experiment}
\subsection{Datasets and Evaluation Metrics}
\textbf{Scene Flow}~\cite{dispNetC} is a synthetic dataset containing 35,454 training pairs and 4,370 testing pairs with dense disparity maps. We use the Finalpass of Scene Flow for training and testing since it is more like real-world images than the Cleanpass, which contains more motion blur and defocus. The end-point error (EPE) and disparity outlier rate Bad 3.0 are used as the evaluation metrics. The Bad 3.0 is defined as the pixels whose disparity errors are larger than 3px.

\textbf{KITTI 2012}~\cite{kitti2012} and \textbf{KITTI 2015}~\cite{kitti2015} are datasets for real-world driving scenes. KITTI 2012 contains 194 training pairs and 195 testing pairs, and KITTI 2015 contains 200 training pairs and 200 testing pairs. Both datasets provide sparse ground-truth disparities obtained with LiDAR.
For KITTI 2012, we report the percentage of pixels with errors larger than x disparities in both non-occluded (x-noc) and all regions (x-all), as well as the overall EPE in both non-occluded (EPE-noc) and all the pixels (EPE-all). For KITTI 2015, we report the percentage of pixels with EPE larger than 3 pixels in background regions (D1-bg), foreground regions (D1-fg), and all (D1-all).

\textbf{Middlebury V3}~\cite{middlebury} is an indoor dataset, which provides 15 training pairs and 15 testing pairs, where some samples are under inconsistent illumination or color conditions. All of the images are available in three different resolutions. We select the full-resolution and half-resolution ones of training pairs to evaluate cross-domain generalization performance on large disparity.
The evaluation metric is bad 2.0, where the percentage of the pixels with EPE is larger than 2 pixels. \textbf{ETH3D}~\cite{eth3d} is a gray-scale dataset with 27 training pairs and 20 testing pairs for indoor and outdoor scenes.

\subsection{Implemention Details}
We implement our IGEV++ with PyTorch and perform our experiments using NVIDIA RTX 3090 GPUs. For all training, we use the AdamW~\cite{adamw} optimizer and clip gradients to the range [-1, 1]. On the Scene Flow dataset, we train IGEV++ for 200k steps with a batch size of 8. We randomly crop images to $256\times768$ and use data augmentation including asymmetric chromatic augmentations and spatial augmentations for training. The indexing radius is set to 4. For all experiments, we use a one-cycle learning rate schedule with a learning rate of 2e-4, and we use 22 update iterations during training. In our experiments, we set $\lambda_{s}$, $\lambda_{m}$, and $\lambda_{l}$ to 1.0, 0.5, and 0.2, respectively. $\gamma$ is set to 0.9.

\begin{table*}
    \centering
    \caption{Quantitative comparisons on the KITTI 2012~\cite{kitti2012} and KITTI 2015~\cite{kitti2015} test sets. We categorize state-of-the-art methods into two groups based on their runtime and further into cost volume filtering-based ($\textit{Filter.}$), transformer-based ($\textit{Trans.}$), and iterative optimization-based ($\textit{Iter.}$) methods. $^\dag$ indicates that we used DepthAnythingV2-Large~\cite{depthanyv2} was used as the feature extraction backbone, with frozen parameters during training. IGEV++ and $^\dag$IGEV++ run 16 update iterations at inference, and RT-IGEV runs 6 update iterations at inference. \textbf{Bold}: Best.}
    \begin{tabular}{l|l|l|cccccc|ccc|c}
     \toprule
     \multirow{2}{*}{Target} & \multirow{2}{*}{Category} & \multirow{2}{*}{Method} & \multicolumn{6}{c|}{KITTI 2012~\cite{kitti2012}} & \multicolumn{3}{c|}{ KITTI 2015~\cite{kitti2015}} & \multirow{2}{*}{\makecell{Run-time\\(ms)}} \\
     & & & 2-noc & 2-all & 3-noc & 3-all & \thead{EPE \\ noc} & \thead{EPE\\all} & D1-bg & D1-fg & D1-all & \\
     \midrule

    \multirow{8}{*}{ \rotatebox{90}{\textit{Real-Time}}} & \multirow{7}{*}{\textit{Filter.}} & HITNet~\cite{hitnet} & 2.00 & 2.65 &1.41  & 1.89 &  0.4 & 0.5 & {1.74}  & \textbf{3.20} & {1.98} & 20\\
    & & DeepPrunerFast~\cite{deeppruner} & - & - & - & - & - & - & 2.32 & 3.91 & 2.59 & 50\\
     & & AANet~\cite{aanet} & 2.30 & 2.96  & 1.55 &2.05 & 0.4 & 0.5 &  1.65 & 3.96 & 2.03 & 60\\
    & & BGNet+~\cite{bgnet}  & 2.78 & 3.35 & 1.62 & 2.03 & 0.5 & 0.6 & 1.81 & 4.09  & 2.19 & 35\\
    & & {DecNet~\cite{decomposition}}  & - & - & - & - & - & - & 2.07 & 3.87 & 2.37 & 50\\
    & & CoEx~\cite{coex} & 2.54 & 3.09 &  1.55 & 1.93 & 0.5 & 0.5 & 1.74 & 3.41  & 2.02 & 27\\
    &  & Fast-ACVNet+~\cite{fast-acv} & 2.39 & 2.97 & 1.45 & 1.85 & 0.5 & 0.5 & 1.70 & 3.53  & 2.01 & 45\\
    \cline{2-13}
    \rule{0pt}{10pt}
    & \textit{Iter.} & RT-IGEV (Ours) & \textbf{1.93} & \textbf{2.51} & \textbf{1.29} & \textbf{1.68} & 0.4 & 0.5  & \textbf{1.48} & 3.37 & \textbf{1.79} & 48 \\
    \midrule
    
    \multirow{22}{*}{\rotatebox{90}{\textit{Accuracy}}} & \multirow{11}{*}{\textit{Filter.}} & PSMNet~\cite{psmnet}  & {2.44} & {3.01} &1.49 & 1.89 & 0.5 & 0.6 & 1.86 & 4.62 & 2.32 & 410 \\
    & & GwcNet~\cite{gwcnet} & 2.16 & 2.71 & 1.32 & 1.70 & 0.5 &0.5 & 1.74 & 3.93 & 2.11 & 320 \\
    & & GANet-deep~\cite{ganet}  & 1.89 & 2.50 & 1.19 & 1.60 & 0.4 &0.5 & 1.48 & 3.46 & 1.81 & 1800 \\
    & & AcfNet~\cite{acfnet} & 1.83 & 2.35 &{1.17}  & {1.54} &  {0.5} & {0.5} & {1.51}  & {3.80} & {1.89} & 480\\
    & & CFNet~\cite{cfnet} & {1.90} & {2.43} &{1.23}  & {1.58} &0.4 & {0.5} & {1.54}  & {3.56} & {1.88} & 180\\
    & & EdgeStereo-V2~\cite{edgestereo} &2.32 &2.88 &1.46 &1.83 & 0.4 & 0.5 &1.84 &3.30 & 2.08 & 320 \\
    & & CSPN~\cite{cfnet} & {1.79} & {2.27} &{1.19}  & {1.53} &- &- & {1.51}  & {2.88} & {1.74} & 1000\\
    & & LEAStereo~\cite{leastereo}  & 1.90 & 2.39 & 1.13 & 1.45 & 0.5 &0.5 & 1.40 & 2.91 & 1.65 & 300 \\
    & & ACVNet~\cite{acvnet} & 1.83 & 2.35 & 1.13 & 1.47 & 0.4 &0.5 & 1.37 & 3.07 & 1.65 & 200 \\
    & & PCWNet~\cite{pcwnet} & 1.69 & 2.18 & 1.04 & 1.37 & 0.4 & 0.5 & 1.37 & 3.16 & 1.67 & 440 \\
    & & GANet+ADL~\cite{ADL} & 1.52 & 2.01 & 0.98 & 1.29 & 0.4 & 0.5  & 1.38 & 2.38 & 1.55 & 670\\
    \cline{2-13}
    \rule{0pt}{10pt}
    & \multirow{2}{*}{\textit{Trans.}} & GMStereo~\cite{unistereo} &- & - & - & - & - & - & 1.49 & 3.14 & 1.77 & 170 \\
    & & Croco-Stereo~\cite{croco-stereo} & - & - & - & - & - & - & 1.38 & 2.65 & 1.59 & 930 \\
    \cline{2-13}
    \rule{0pt}{10pt}
    & \multirow{9}{*}{\textit{Iter.}} & RAFT-Stereo~\cite{raft-stereo} &1.92 &2.42 &1.30 & 1.66 & 0.4 & 0.5 &1.58 &3.05 &1.82 & 380 \\ 
    & & DLNR~\cite{dlnr} & - & - & - & - & - & - & 1.60 & 2.59 & 1.76 & - \\ 
    & & CREStereo~\cite{crestereo} &1.72 &2.18 &1.14 & 1.46 & 0.4 & 0.5 &1.45 &2.86 &1.69 & 410 \\
    & & Selective-IGEV~\cite{selective} & 1.59 & 2.05 & 1.07 & 1.38 & 0.4 & 0.4 & 1.33 & 2.61 & 1.55 & 240 \\
    & & TC-Stereo~\cite{tcstereo} & - & - & - & - & - & - &1.29 &\textbf{2.33} & 1.46 & 90 \\
    & & ViTAStereo~\cite{vitstereo} & 1.46 & 1.80 & 0.93 & 1.16 & 0.4 & 0.4 &1.21 &2.99 & 1.50 & 360 \\
    & & IGEV (Ours) & 1.71 & 2.17 & 1.12 & 1.44 & 0.4 & 0.4 &1.38 &2.67 & 1.59 & 180 \\
    & & IGEV++ (Ours) & 1.56 & 2.03 & 1.04 & 1.36 & 0.4 & 0.4 &1.31 &2.54 & 1.51 & 280 \\
    & & $^\dag$IGEV++ (Ours) & \textbf{1.36} & \textbf{1.74} & \textbf{0.89} & \textbf{1.13} & 0.4 & 0.4 &\textbf{1.15} &2.80 & \textbf{1.43} & 480 \\
    \bottomrule
    \end{tabular}
\label{tab:kitti}
\end{table*}

\subsection{Comparisons with Existing Representative Methods}
We compare our IGEV++ with existing representative methods including GwcNet~\cite{gwcnet}, PCWNet~\cite{pcwnet}, RAFT-Stereo~\cite{raft-stereo}, and GMStereo~\cite{unistereo}, on the Scene Flow dataset across multiple disparity ranges, shown in Table \ref{tab:large_disp}. GwcNet~\cite{gwcnet} and PCWNet~\cite{pcwnet} are cost volume filtering-based methods, RAFT-Stereo~\cite{raft-stereo} is an iterative optimization-based method, and GMStereo~\cite{unistereo} is a transformer-based method. To enable cost volume filtering-based methods (PCWNet~\cite{pcwnet} and GwcNet~\cite{gwcnet}) to handle large disparities, we extend the original pre-defined 192px disparity range to 768px. Our method consistently outperforms these methods by a large margin across all disparity ranges (Table~\ref{tab:large_disp}). Specifically, as the disparity range increases, existing representative methods such as PCWNet~\cite{pcwnet}, RAFT-Stereo~\cite{raft-stereo}, and GMStereo~\cite{unistereo} exhibit a significant drop in accuracy. In contrast, our method remains robust for large disparity ranges. We also compared the memory consumption with other methods (Table~\ref{tab:large_disp}). When handling large disparities, our IGEV++ can consume 7$\times$ less memory than the state-of-the-art volume filtering-based method PCWNet. 

For cost volume filtering-based methods, simply increasing the pre-defined disparity range to construct a large-size cost volume will result in high memory consumption. 
Moreover, a large-size cost volume will significantly increase the difficulty of cost aggregation and disparity regression, leading to a large accuracy degradation. For example, the EPE metric of PCWNet drops from 0.78 to 1.52, a decrease of 94.87\%. In comparison, our IGEV++ constructs multi-range and multi-granularity cost volumes, avoiding the construction of a large-size cost volume, significantly reducing memory consumption and the difficulty of cost aggregation, enabling accurate and efficient handling of large disparities. We display the visual comparisons in Figure~\ref{fig:teaser}. Previous state-of-the-art methods struggle to handle large disparities in large close-range objects; they heavily rely on contextual information, leading to erroneous results. In contrast, our MGEV effectively utilizes multi-granularity matching information, excelling in handling large disparities and textureless regions.

\subsection{Ablation Study}
\textbf{Effectiveness of MGEV.} We explore the best settings for the MGEV and examine its effectiveness. For all models in these experiments, we perform 32 iterations of ConvGRUs update at inference. We take RAFT-Stereo~\cite{raft-stereo} as our baseline, and replace its original backbone with MobileNetV2 100 for efficiency. As shown in Table~\ref{tab:ablation}, the proposed GEV significantly improves prediction accuracy within the 192px disparity range (Disp\textless192px). Furthermore, the proposed MGEV can significantly enhance performance in large disparity regions by a substantial margin. The all-pairs correlations of RAFT-Stereo~\cite{raft-stereo} (Baseline) lack non-local geometry and context information, and thus make it difficult to handle local ambiguities in ill-posed regions. In contrast, our MGEV provides more comprehensive yet concise geometry information to ConvGRUs, yielding more effective optimization in each iteration. For each iteration, we index the fused geometry features from three ranges and three granularities of geometry encoding volumes. Our proposed selective geometry feature fusion module enables the model to better handle disparities of various ranges. The final model, referred to as IGEV++, achieves the best performance across all disparity ranges.

\textbf{Analysis of GEV at different ranges.} To validate the impact of different ranges of GEV on the final disparity results, we separately constructed small-range GEV, medium-range GEV, and large-range GEV, with the iterative optimization operations consistent with those in IGEV++. Accordingly, during the iterative process, the geometry feature $\mathbf{f}_{G}$ is derived from the small-range GEV, medium-range GEV, and large-range GEV, respectively. As can be seen from Table~\ref{tab:ablation_sml} and Figure~\ref{fig:comp_sml}, the iterative small-range GEV performs well for small disparities but struggles with large disparities. On the other hand, the iterative large-range GEV handles large disparities effectively; however, due to coarse-grained matching, it tends to lose details in small-range disparity regions. Our MGEV combines the strengths of both, achieving the best performance.

\begin{figure*}
\centering
\includegraphics[width=0.9\textwidth]{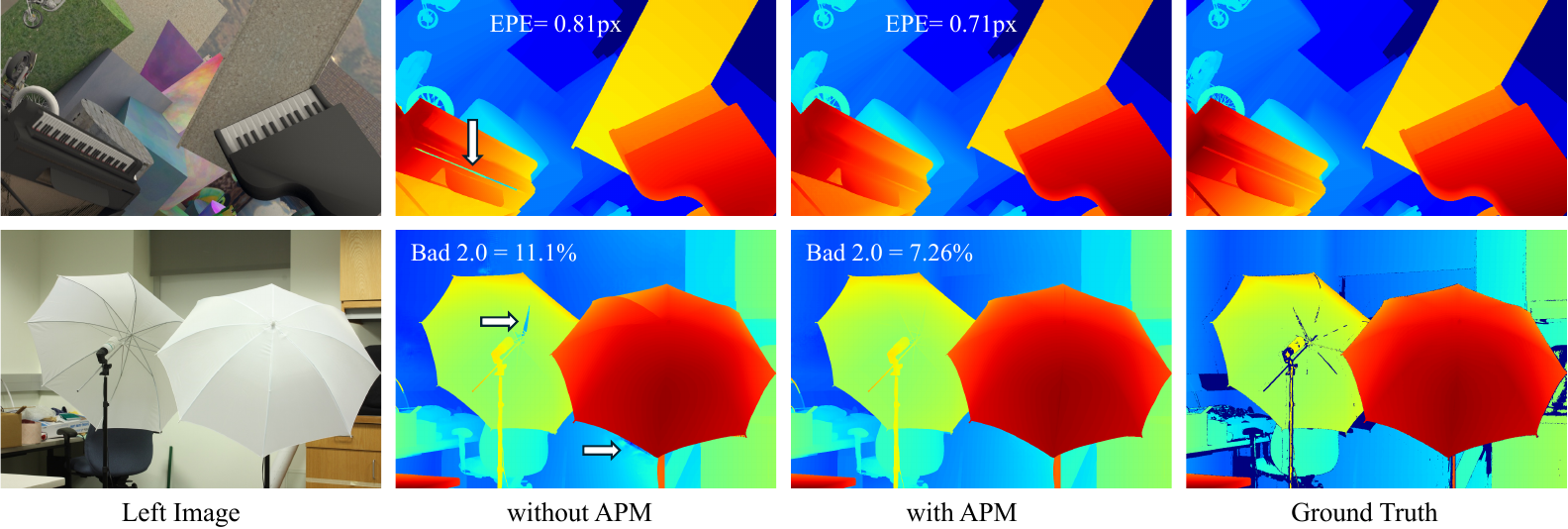} 
\vspace{-5pt}
\caption{Qualitative comparisons with/without APM module. Row 1 shows performance on the Scene Flow test set, while Row 2 presents the zero-shot generalization performance on the Middlebury dataset. Our APM is more robust in handling large ill-posed regions and large disparities.}
\label{fig:comp_apm}
\end{figure*}

\textbf{Adaptive Patch Matching.} When adaptive patch matching (APM) is not used, MGEV involves constructing multi-size cost volumes at 1/4 resolution. Specifically, the large-size cost volume contains 192 disparity candidates (768/4) for cost aggregation, the medium-size cost volume contains 96 disparity candidates (384/4), and the small-size cost volume contains 48 disparity candidates (192/4) for cost aggregation. The use of APM enables the efficient construction of MGEV, where only 48 disparity candidates (768/4/4) are used for cost aggregation in the large-range GEV, as well as in the medium disparity range GEV. Moreover, a large-size (192 disparity candidates) cost volume significantly increases the complexity of cost aggregation and disparity regression, particularly in extensive textureless regions and areas with large disparities, as shown in Figure~\ref{fig:comp_apm} and Table~\ref{tab:ablation}.

\begin{figure*}
\centering
\includegraphics[width=0.8\textwidth]{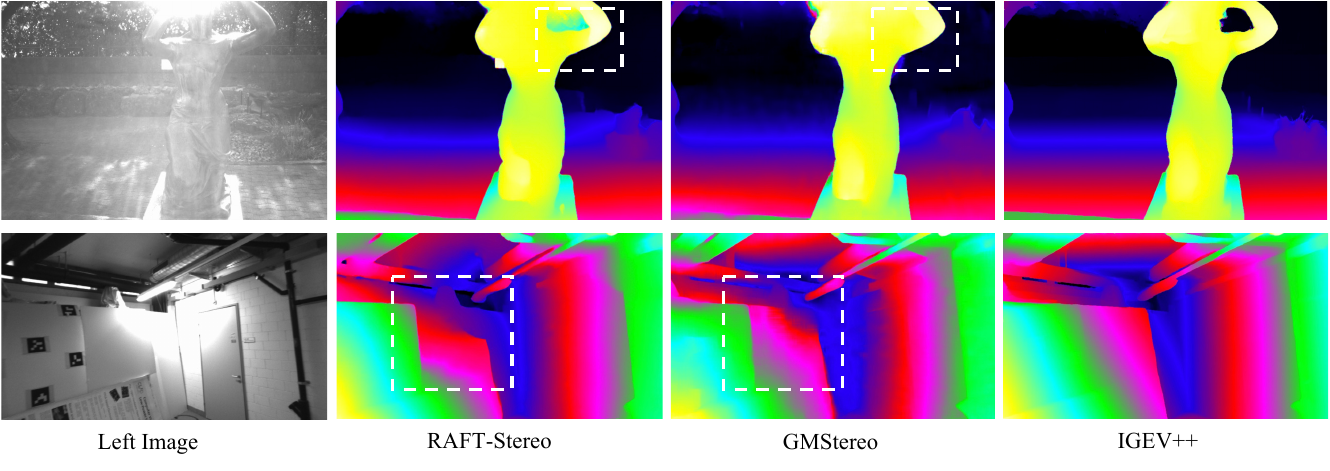} 
\vspace{-10pt}
\caption{Visual comparisons with state-of-the-art stereo methods~\cite{raft-stereo,unistereo} on the ETH3D test set~\cite{eth3d}.}
\label{fig:eth3d}
\end{figure*}

\textbf{Number of Iterations}.
Our IGEV++ achieves excellent performance even when the number of iterations is reduced. As shown in Table~\ref{tab:iter}, we report the EPE (Disp\textless768px) of RAFT-Stereo~\cite{raft-stereo}, DLNR~\cite{dlnr} and our IGEV++ on the Scene Flow test set~\cite{dispNetC}. Compared with all-pairs correlations in RAFT-Stereo and DLNR, our MGEV can facilitate more comprehensive yet concise geometry information to ConvGRUs, yielding more effective optimization in each iteration and thus effectively addressing local ambiguities in ill-posed regions. Thus when the number of iterations is reduced to 1, 2, 3 or 4, our IGEV++ outperforms RAFT-Stereo~\cite{raft-stereo} and DLNR~\cite{dlnr} with the same number of iterations by a large margin, e.g., surpassing RAFT-Stereo and DLNR by 72.91\% and 57.27\%, respectively, when the iteration number is 1. From Table~\ref{tab:iter} and Figure~\ref{fig:iterations}, we can observe that our IGEV++ achieves state-of-the-art performance even with few iterations, providing users a flexible trade-off between time efficiency and accuracy according to their needs.

\subsection{Benchmark Results}
We perform system-level comparisons with previous methods on the KITTI 2012~\cite{kitti2012}, KITTI 2015~\cite{kitti2015}, Middlebury~\cite{middlebury}, and ETH3D~\cite{eth3d} benchmarks. 

\textbf{KITTI}. The comparison results with previous methods are shown in Table~\ref{tab:kitti}.  Our IGEV++ achieves the best performance among the published methods for almost all metrics on KITTI 2012~\cite{kitti2012} and 2015~\cite{kitti2015}. We first fine-tune the pre-trained Scene Flow~\cite{dispNetC} model on the virtual KITTI 2~\cite{vkitti} for 60k steps with a batch size of 8. Then we fine-tune it on the mixed dataset of KITTI 2012~\cite{kitti2012} and KITTI 2015~\cite{kitti2015} for 50k steps. The quantitative comparison results are shown in Table~\ref{tab:kitti}. On KITTI 2012, our IGEV++ outperforms LEAStereo~\cite{leastereo} and CREStereo~\cite{crestereo} by 17.89\% and 9.30\% on 2-noc metric, respectively. On KITTI 2015, our IGEV++ surpasses GMStereo~\cite{unistereo} and Croco-Stereo~\cite{croco-stereo} by 14.69\% and 9.58\% on D1-all metric, respectively. As shown in Table IV, IGEV++ increases the runtime by 100ms compared to IGEV. Prior to iterative optimization, IGEV++ incurs an additional 40ms for constructing and filtering the medium-range and large-range cost volumes, whereas IGEV only constructs and filters a single small-range cost volume. During the iterative process (16 iterations during inference), IGEV++ further introduces overhead by indexing a set of geometry features $\mathbf{f}_{G}^{m}$  and $\mathbf{f}_{G}^{l}$ from the medium-range GEV and large-range GEV at each iteration, and by computing selective weights to fuse these geometry features. This step contributes an additional 60ms to the overall runtime.

\begin{table}
  \centering
  \caption{
  Quantitative results on the Scene Flow test set for different iterations. The evaluation metric is EPE (Disp\textless768px). Our IGEV++ achieves better results than RAFT-Stereo with only 2 iterations compared to its 32 iterations.}
  \begin{tabular}{lccccccc}
    \toprule
    \multirow{2}{*}{Model} &\multicolumn{7}{c}{Number of Iterations}\\
    & 1 & 2 & 3 & 4 & 8 & 16 & 32 \\
    \midrule
    RAFT-Stereo~\cite{raft-stereo} & 3.58 & 2.08 & 1.64 & 1.42 & 1.11 & 1.00 & 0.98\\
    DLNR~\cite{dlnr}  & 2.27 & 1.42 & 1.21 & 1.09 & 0.90 & 0.82 & 0.81\\
    IGEV++  &\textbf{0.97} &\textbf{0.88} &\textbf{0.83} &\textbf{0.79} &\textbf{0.71} &\textbf{0.67} & \textbf{0.67}\\
    \bottomrule
  \end{tabular}
  \label{tab:iter}
\end{table}

\begin{table}
  \centering
\caption{Evaluation in the reflective regions (ill-posed regions) of KITTI 2012 benchmark. } 
  \begin{tabular}{l|cccc}
    \toprule
    \multirow{2}{*}{Method} & \multicolumn{4}{c}{KITTI 2012 (Reflective)} \\
    & 2-noc & 2-all &3-noc &3-all \\
    \midrule
    RAFT-Stereo~\cite{raft-stereo} & 8.41 & 9.87 & 5.40 & 6.48 \\
    Any-RAFT~\cite{any-raft} & 7.79  & 9.15 &4.85 & 5.71 \\
    PCWNet~\cite{pcwnet} & 8.94 & 10.71 & 4.99 & 6.20 \\
    RiskMin~\cite{riskmin} & 7.57 & 9.60 & 4.11  & 5.51 \\
    GANet+ADL~\cite{ADL} & 8.57 & 10.42 & 4.84 & 6.10 \\
    $\text{HD}^3$-Stereo~\cite{hd3} & 8.61 & 10.89 & 4.99 & 6.77 \\
    LEAStereo~\cite{leastereo} & 9.66 & 11.40 & 5.35 & 6.50\\
    CREStereo~\cite{crestereo} & 9.71 & 11.26 & 6.27 & 7.27 \\
    NMRF-Stereo~\cite{nmrf} & 10.02 & 12.34 & 6.35 & 8.11 \\
    UCFNet~\cite{ucfnet} & 9.78 & 11.67 & 5.83 & 7.12\\
    Selective-IGEV~\cite{selective} & 6.73 & 7.84 & 3.79 & 4.38 \\
    GANet+ADL~\cite{ADL} & 8.57 & 10.42 & 4.84 & 6.10 \\
    IGEV++ & \textbf{6.47} & \textbf{7.64} & \textbf{3.71} & \textbf{4.35} \\
    \bottomrule
  \end{tabular}
  \label{tab:reflective}
  \vspace{-5pt}
\end{table}
\begin{table*}[t]
\centering
\caption{Quantitative comparisons with previous methods on ETH3D~\cite{eth3d} and Middlebury~\cite{middlebury} benchmarks. \textbf{Bold}: Best, \underline{Underline}: Second best.}
\begin{tabular}{l|cccc|cccc}
\toprule
\multirow{2}{*}{Method} & \multicolumn{4}{c|}{ETH3D} & \multicolumn{4}{c}{Middlebury} \\
 & Bad 0.5 & Bad 1.0 & Bad 4.0 & AvgErr & Bad 1.0 & Bad 2.0 & Bad 4.0 & AvgErr \\
 \midrule
CroCo-Stereo~\cite{croco-stereo}& {3.27} & \underline{0.99} & 0.13 & 0.14 & 16.9& 7.29  & 4.18 & 1.76 \\
GMStereo~\cite{unistereo} & 5.94 & 1.83 & {0.08} & 0.19 & 23.6 & 7.14  & 2.96 & 1.31\\
HITNet~\cite{hitnet} & 7.83 & 2.79 & 0.19 & 0.20 & 13.3 & 6.46  & 3.81 & 1.71  \\
RAFT-Stereo~\cite{raft-stereo}  & 7.04 & 2.44 & 0.15 & 0.18 & 9.37 & 4.74  & 2.75 & 1.27 \\
CREStereo~\cite{crestereo} & 3.58 & \textbf{0.98} & 0.10 & \underline{0.13} & 8.25 & 3.71  & 2.04 & 1.15 \\
EAI-Stereo~\cite{eai} & 5.21 & 2.31 & 0.70 & 0.21 & 7.81 & 3.68 & 2.14 & 1.09 \\
DLNR~\cite{dlnr} & -  & - & - & - & \underline{6.82} & \underline{3.20}  & {1.89} & {1.06} \\
Selective-RAFT~\cite{selective} & 5.78 & 1.69 & 0.13 & 0.17  & - & - & - & - \\
Selective-IGEV~\cite{selective} & \underline{3.06}  & 1.23 & \textbf{0.05} & \textbf{0.12} & \textbf{6.53} & \textbf{2.54}  & \textbf{1.36} & \textbf{0.91} \\
IGEV++ (Ours) & \textbf{2.98} & 1.14 & \underline{0.06} & \underline{0.13} & {7.18}  & {3.23}   & \underline{1.82} & \underline{0.97} \\
\bottomrule
\end{tabular}
\label{tab:eth3d-middlebury}
\end{table*}

\textbf{Middlebury}. Following CREStereo~\cite{crestereo} and GMStereo~\cite{unistereo}, we first fine-tune the pre-trained Scene Flow model on the mixed Tartan Air~\cite{wang2020tartanair}, CREStereo dataset~\cite{crestereo}, Scene Flow~\cite{dispNetC}, Falling Things~\cite{tremblay2018falling}, InStereo2k~\cite{bao2020instereo2k}, CARLA HR-VS~\cite{yang2019hierarchical} and Middlebury~\cite{middlebury} datasets using a crop size of $384 \times 512$ for 200k steps. Then we fine-tune it on the mixed CREStereo dataset~\cite{crestereo}, Falling Things~\cite{tremblay2018falling}, InStereo2k~\cite{bao2020instereo2k}, CARLA HR-VS~\cite{yang2019hierarchical} and Middlebury~\cite{middlebury} datasets using a crop size of $384 \times 768$ with a batch size of 8 for another 100k steps. As shown in Table. \ref{tab:eth3d-middlebury}, our IGEV++ outperforms most published methods on Middlebury~\cite{middlebury} benchmark. Visual comparisons are shown in Figure~\ref{fig:middlebury}. 

\textbf{ETH3D.} Following CREStereo~\cite{crestereo} and GMStereo~\cite{unistereo}, we use a collection of several public stereo datasets for training.  We first fine-tune the pre-trained Scene Flow model on the mixed Tartan Air~\cite{wang2020tartanair}, CREStereo dataset~\cite{crestereo}, Scene Flow~\cite{dispNetC}, Sintel Stereo~\cite{butler2012naturalistic}, InStereo2k~\cite{bao2020instereo2k} and ETH3D~\cite{schops2017multi} datasets for 300k steps, and the crop size is $384 \times 512$. Then we fine-tune it on the mixed CREStereo dataset~\cite{crestereo}, InStereo2k~\cite{bao2020instereo2k} and ETH3D~\cite{eth3d} datasets for another 100k steps. As shown in Table \ref{tab:eth3d-middlebury}, our IGEV++ outperforms most published methods on ETH3D~\cite{eth3d} benchmark. Visual comparisons are shown in Figure~\ref{fig:eth3d}, our IGEV++ performs better than RAFT-Stereo~\cite{raft-stereo} and GMStereo~\cite{unistereo} in textureless regions.

\begin{table} 
\begin{center}
\caption{Comparisons with other real-time methods on the Scene Flow test set~\cite{dispNetC}} \label{tab:real_time_sf}
\begin{tabular}{lcc}
\toprule
{Method} & EPE (Disp\textless192px) & Run-time (ms) \\ 
\midrule
DeepPrunerFast~\cite{deeppruner} & 0.97 & 61 \\
AANet~\cite{aanet} & 0.87 & 62 \\
BGNet~\cite{bgnet} & 1.17 & 28 \\
DecNet~\cite{decomposition} & 0.84 & 50 \\
CoEx~\cite{coex} & 0.69 & 33 \\
Fast-ACVNet+~\cite{fast-acv} & 0.59 & 45 \\
RT-IGEV (Ours, Iters=2) & 0.60 & 36 \\
RT-IGEV (Ours, Iters=4) & 0.55 & 42 \\
RT-IGEV (Ours, Iters=6) & 0.52 & 48 \\
RT-IGEV (Ours, Iters=8) & \textbf{0.50} & 54 \\
\bottomrule
\end{tabular}
\end{center}
\end{table}

\begin{figure}
\centering
{\includegraphics[width=1.0\linewidth]{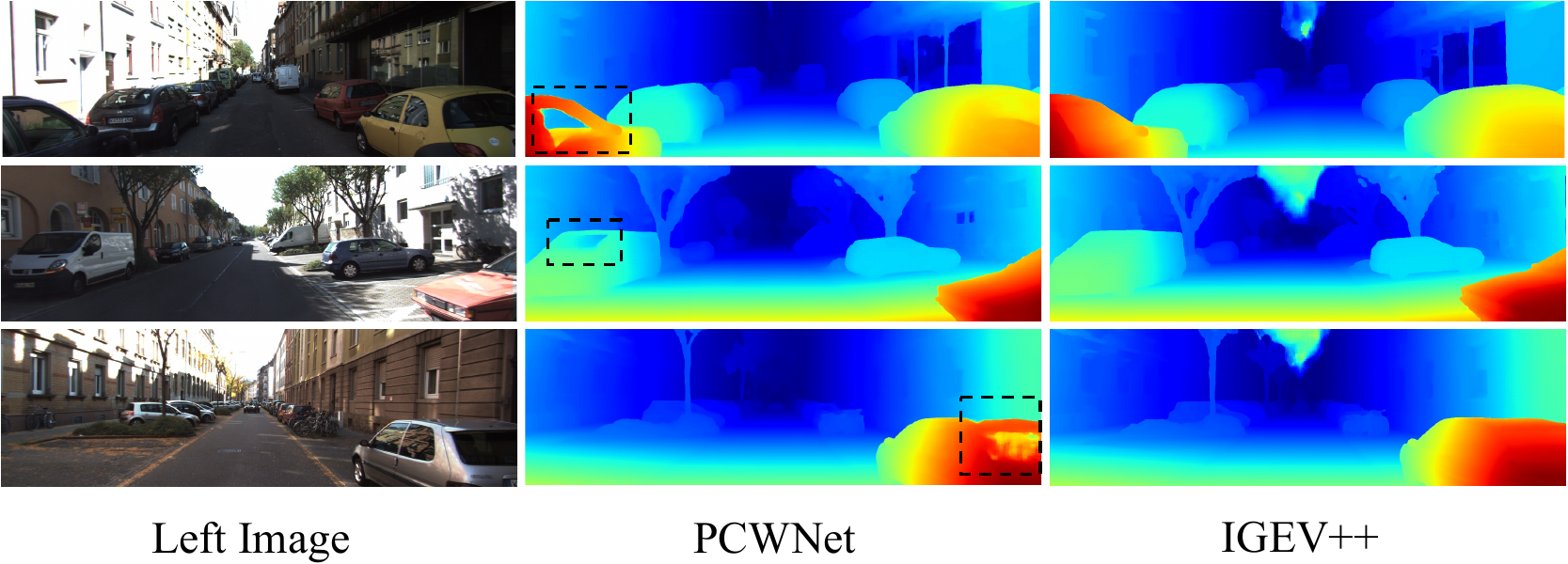}}
\caption{Qualitative comparisons on the test set of KITTI 2012~\cite{kitti2012}. Our IGEV++ performs very well in reflective regions (ill-posed regions).}\label{fig:kt12}
\vspace{-5pt}
\end{figure}

\begin{table}
  \centering
  \caption{
   Zero-shot generalization results on the large disparity dataset Middlebury V3. All models are trained on the synthetic Scene Flow dataset and then tested directly on the real Middlebury V3. The evaluation metric is the 2-pixel outlier rate (Bad 2.0).}
  \begin{tabular}{lccc}
    \toprule
    \multirow{2}{*}{Model} &\multicolumn{2}{c}{Middlebury} \\
    & half & full & \\
    \midrule
    PSMNet~\cite{psmnet} & 25.1 & 39.5\\
    GANet~\cite{ganet} & 20.3 & 32.2 \\
    CFNet~\cite{cfnet} &15.3 & 28.2 \\
    PCWNet~\cite{pcwnet} & 14.4 & - \\
    DiffuVolume~\cite{diffuvolume} & 10.8 & 15.2 \\
    RAFT-Stereo~\cite{raft-stereo} & 12.6 & 18.3 \\
    GMStereo~\cite{unistereo} & 18.2 & - \\
    DLNR~\cite{dlnr} & 9.5 & 14.5 \\
    IGEV++ (Ours) & \textbf{7.8} & \textbf{12.7} \\
    \bottomrule
  \end{tabular}
  \label{tab:generalization}
  \vspace{-5pt}
\end{table}

\subsection{Performance in Ill-posed Regions}
To verify the performance of our IGEV++ in ill-posed regions, we compare our method with previously published state-of-the-art methods on reflective regions in the KITTI 2012 benchmark. As shown in Table~\ref{tab:reflective}, our IGEV++ achieves the best performance among all methods. Compared to the all-pairs correlations of RAFT-Stereo, our MGEV encodes sufficient non-local geometry information and scene prior knowledge, which are essential for matching in ill-posed regions. Thus, our IGEV++ outperforms RAFT-Stereo~\cite{raft-stereo} by 31.30\% on the KITTI 2012 reflective regions (Table~\ref{tab:reflective}) on the 3-noc metric. Visual comparisons are shown in Figure~\ref{fig:kt12} that our IGEV++ performs very well in the reflective regions.

\subsection{Real-Time Version of IGEV++}
To meet the requirement of time-constrained applications, we further demonstrate that the proposed IGEV++ can be easily configured to achieve real-time inference on KITTI-resolution ($1248\times384$) images. Compared to IGEV++, the real-time version (named RT-IGEV), is based on four main modifications. First, we remove the context network and use a single backbone (feature network) to generate context features. Second, we construct a single-range GEV with a pre-defined maximum disparity of 192px instead of MGEV. Third, since our GEV already integrates contextual cues and contains non-global geometry information, we use a single-level ConvGRU to replace RAFT-Stereo's three-level ConvGRUs. Finally, for efficiency, we reduce the hidden state channels of the ConvGRU from 128 to 96. As shown in Table~\ref{tab:real_time_sf}, our RT-IGEV achieves the best performance among all published methods on the Scene Flow test set. We present results for iteration numbers of 2, 4, 6, and 8, enabling users to trade-off between time efficiency and performance according to their needs. We also evaluate RT-IGEV on the KITTI 2012 and 2015 test sets, the results are shown in Table~\ref{tab:kitti}. Our RT-IGEV ranks first among all published real-time methods. On KITTI 2012, RT-IGEV outperforms HITNet~\cite{hitnet} and Fast-ACVNet+~\cite{fast-acv} by 8.51\% and 11.03\% on 3-noc metric, respectively. On KITTI 2015, RT-IGEV surpasses HITNet~\cite{hitnet} and Fast-ACVNet+~\cite{fast-acv} by 13.64\% and 10.95\% on D1-all metric, respectively.

\subsection{Zero-shot Generalization Performance}
Since large-scale real-world datasets for training models are difficult to obtain, the zero-shot generalization ability of stereo models is crucial. We evaluate the zero-shot generalization performance of IGEV++ from synthetic Scene Flow datasets to unseen real-world scenes. In this evaluation, we train our IGEV++ on the Scene Flow dataset~\cite{dispNetC}, and then directly test it on the Middlebury V3 training sets. As shown in Table~\ref{tab:generalization}, our IGEV++ achieves state-of-the-art performance in the zero-shot setting. Figure~\ref{fig:teaser} shows the visual comparisons with PCWNet~\cite{pcwnet}, DLNR~\cite{dlnr}, GMStereo~\cite{unistereo}, our IGEV++ effectively distinguishes subtle details in complex backgrounds. 

\begin{figure}
\centering
{\includegraphics[width=0.9\linewidth]{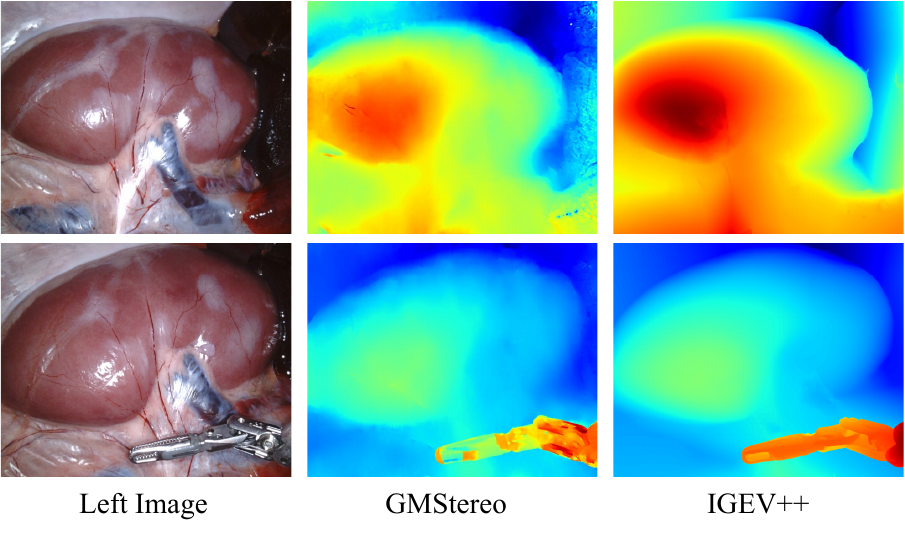}}
\vspace{-5pt}
\caption{Zero-shot generalization results on the large disparity dataset SCARED~\cite{scared}. All models are trained solely on the synthetic Scene Flow dataset. Our IGEV++ performs well in textureless regions and can predict sharp edges.}\label{fig:scared}
\end{figure}

In medical applications~\cite{liu2024endogaussian,wang2022neural,yang2023neural,ma2024segment,chen2023confidence,chen2021transunet} like endoscopic intervention, recognizing close objects with large disparities is crucial to avoid accidental damage to patient tissues or organs. We also evaluate the generalization performance on the large disparity SCARED~\cite{scared} dataset. As shown in Figure~\ref{fig:scared}, our IGEV++ outperforms the state-of-the-art transformer-based method GMStereo~\cite{unistereo}, excelling in textureless regions and accurately predicting clear edges of tissues or organs.

\section{Conclusion}
This paper presents IGEV++, a novel stereo matching network architecture that takes full advantage of both filtering-based and iterative optimization-based methods while overcoming their respective limitations. Specifically, IGEV++ builds a geometry encoding volume that integrates spatial cues and encodes geometry information, and then iteratively indexes it to update the disparity map. 

To efficiently handle large disparities and textureless/reflective regions, we further propose multi-range geometry encoding volumes (MGEV), which encode coarse-grained geometry information for ill-posed
regions and large disparities, while preserving fine-grained geometry information for details and small disparities.  To effectively and efficiently construct MGEV and fuse geometry features across multiple ranges and granularities within MGEV, we introduce an adaptive patch matching module and a selective geometry feature fusion module, respectively. 

Our IGEV++ achieves the best performance on the Scene Flow test set across all disparity ranges, up to 768px. Our IGEV++ also achieves state-of-the-art accuracy on the Middlebury, ETH3D, KITTI 2012 and 2015 benchmarks, and exhibits impressive generalization ability to unseen real-world datasets.

We also propose RT-IGEV, a real-time version of IGEV++, which achieves real-time inference while delivering the best performance among all published real-time methods.

\section*{Acknowledgments}
This work is supported by the National Natural Science Foundation of China (623B2036, 62472184) and the Fundamental Research Funds for the Central Universities.

\bibliographystyle{IEEEtran}
\bibliography{IGEV++}






 

\vspace{40pt}
\begin{IEEEbiography}[{\includegraphics[width=1in,height=1in,clip,keepaspectratio]{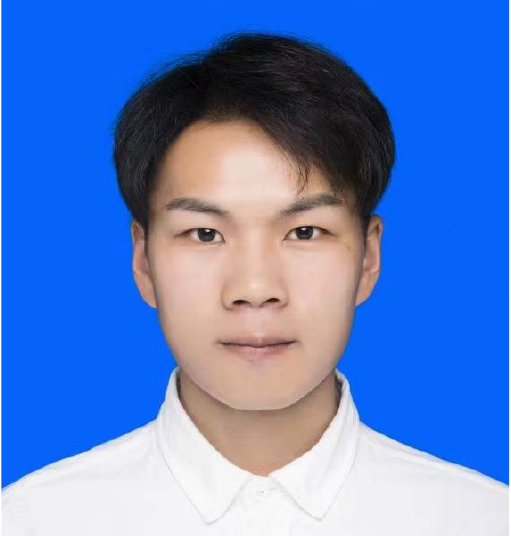}}]{Gangwei Xu}
  is a PhD student at the Department of Electronic Information and Communications at Huazhong University of Science and Technology. He is supervised by Prof. Xin Yang. He received his B.Eng. degree from Huazhong University of Science and Technology in 2021. His research interests include stereo matching, optical flow estimation and HDR video reconstruction. He has published multiple papers in IEEE-TPAMI, IJCV, and CVPR.
\end{IEEEbiography}
\vspace{-15pt}

\vspace{-15pt}
\begin{IEEEbiography}[{\includegraphics[width=1in,height=1in,clip,keepaspectratio]{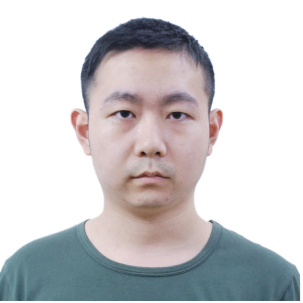}}]{Xianqi Wang}
  is a PhD student at the Department of Artificial Intelligence and Automation at Huazhong University of Science and Technology. He is supervised by Prof. Xin Yang. He received his B.Eng. degree from Huazhong University of Science and Technology in 2022. His research interests include stereo matching and multi-view stereo. 
\end{IEEEbiography}
\vspace{-15pt}

\vspace{-15pt}
\begin{IEEEbiography}[{\includegraphics[width=1in,height=1in,clip,keepaspectratio]{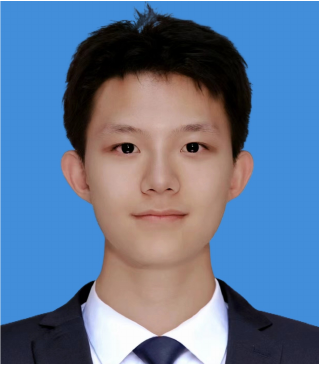}}]{Zhaoxing Zhang}
  is a master's student at the School of Electronic Information and Communications at Huazhong University of Science and Technology. He is supervised by Prof. Xin Yang. He received his B.Eng. degree from Huazhong University of Science and Technology in 2024. His research focuses on visual odometry, stereo matching, and optical flow estimation. 
\end{IEEEbiography}
\vspace{-15pt}

\vspace{-15pt}
\begin{IEEEbiography}[{\includegraphics[width=1in,height=1in,clip,keepaspectratio]{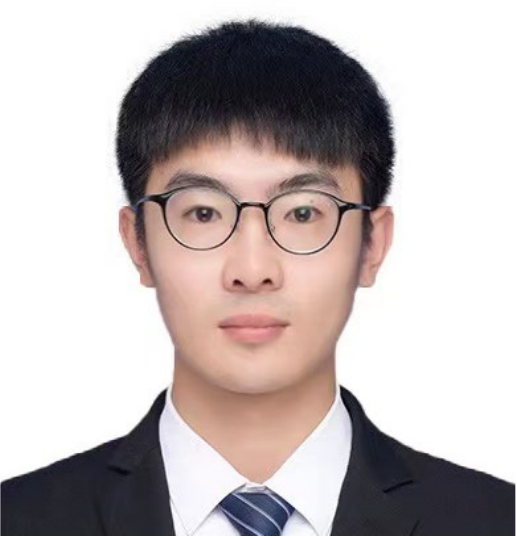}}]{Junda Cheng}
  is a PhD student at the Department of Electronic Information and Communications at Huazhong University of Science and Technology. He is supervised by Prof. Xin Yang. He received his B.Eng. degree from Huazhong University of Science and Technology in 2020. His research interests include stereo matching,  multi-view stereo, and deep visual odometry. 
\end{IEEEbiography}
\vspace{-15pt}

\vspace{-15pt}
\begin{IEEEbiography}[{\includegraphics[width=1in,height=1in,clip,keepaspectratio]{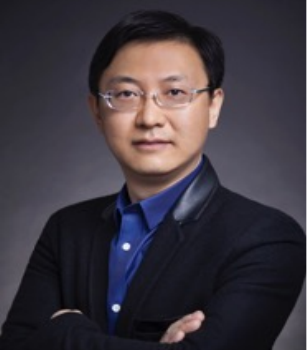}}]{Chunyuan Liao}
  received his PhD degree in University of Maryland, College Park in 2008. He was a research scientist at Fuji Xerox Palo Alto Laboratory in Silicon Valley, where he was granted significant achievement awards three times and ACM conference awards twice due to his outstanding research in Augmented Reality. In the year of 2012, Chunyuan Liao founded HiScene, a leading Chinese Augmented Reality technology provider. 
  Dr. Liao
has published over 60 technical papers, including ACM Trans. on CHI, UIST,
ICCV, ACM MM, CHI, etc., and holds 10+ U.S. Patents.
\end{IEEEbiography}
\vspace{-15pt}

\vspace{-10pt}
\begin{IEEEbiography}[{\includegraphics[width=1in,height=1in,clip,keepaspectratio]{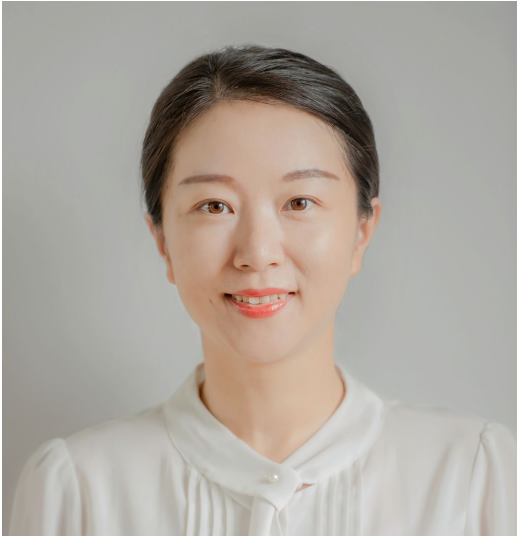}}]{Xin Yang}
   is a Professor at the Department of Electronic Information and Communications at Huazhong University of Science and Technology. She received her Ph.D. degree in the Department of Electrical Computer Engineering at the University of California, Santa Barbara (UCSB).  Her research interests include medical image analysis and 3D vision. She is the recipient of the National Natural Science Fund of China for Excellent Youth Scholar and China Society of Image and Graphics Qingyun Shi Female Scientist Award. She has published over 90 technical papers and held 20 patents. She serves as an Associate Editor of IEEE-TVCG, IEEE-TMI and Multimedia System, an Area Chair of CVPR’24, MICCAI’19-21, and ACM MM’18. She is also a reviewer of top journals such as IEEE-TPAMI, IJCV, etc.
\end{IEEEbiography}


\vfill

\end{document}